\documentclass[10pt,twocolumn,letterpaper]{article}

\usepackage{cvpr}
\usepackage{times}
\usepackage{epsfig}
\usepackage{graphicx}
\usepackage{amsmath}
\usepackage{amssymb}
\usepackage{subfigure}
\usepackage{color}
\usepackage{setspace}
\usepackage{algorithm}
\usepackage{algorithmic}
\usepackage{multirow}

\usepackage[pagebackref=true,breaklinks=true,letterpaper=true,colorlinks,bookmarks=false]{hyperref}

\cvprfinalcopy 

\def\cvprPaperID{1736} 

\ifcvprfinal\fi
\begin{document}

\title{Learning Discriminative Features via Label Consistent Neural Network}


\author{Zhuolin Jiang$^{\dag}$\thanks{Indicates equal contributions.}, Yaming Wang$^{\ddag}$\footnotemark[1], Larry Davis$^\ddag$, Walt Andrews$^\dag$, Viktor Rozgic$^\dag$ \\
$^\dag$Raytheon BBN Technologies, Cambridge, MA, 02138 \\
$^\ddag$University of Maryland, College Park, MD, 20742 \\
{\tt\small \{zjiang,wandrews,vrozgic\}@bbn.com, \{wym,lsd\}@umiacs.umd.edu}
}

\maketitle

\begin{abstract}
Deep Convolutional Neural Networks (CNN) enforce supervised information only at the output layer, and hidden layers
are trained by back propagating the prediction error from the output layer without explicit supervision. We propose a
supervised feature learning approach, Label Consistent Neural Network, which enforces direct supervision in late hidden layers in a novel way.
We associate each neuron in a hidden layer with a particular class label and encourage it to be activated for input signals from the same class.
More specifically, we introduce a label consistency regularization called ``discriminative representation error'' loss
for late hidden layers and combine it with classification error loss to build our overall objective function.
This label consistency constraint alleviates the common problem of gradient vanishing and tends to faster
convergence; it also makes the features derived from late hidden layers discriminative enough for classification even
using a simple $k$-NN classifier, since input signals from the same class will have very similar
representations. Experimental results demonstrate that our approach achieves state-of-the-art performances on several public
benchmarks for action and object category recognition.
\end{abstract}

\section{Introduction}\label{sec1}

Convolutional neural networks (CNN) \cite{lecun98} have exhibited impressive performances in
many computer vision tasks such as
image classification \cite{hinton12}, object detection \cite{rcnn} and image retrieval \cite{cnnretrieval}. When large amounts of training data are available, CNN can automatically learn hierarchical feature
representations, which are more discriminative than previous hand-crafted ones \cite{hinton12}.

Encouraged by their impressive performance in static image analysis tasks, several CNN-based approaches have been developed for action
recognition in videos \cite{3dcnn,karpathy14,joe15,twostream1,trajpool,shengxin15}. Although promising results have been reported,
the advantages of CNN approaches over traditional ones \cite{idt} are not as overwhelming for videos as in static images.
Compared to static images, videos have larger variations in appearance as well as high complexity
introduced by temporal evolution, which makes learning features for recognition from videos more challenging. On the
other hand, unlike large-scale and diverse static image data \cite{imagenet}, annotated data
for action recognition tasks is usually insufficient, since annotating massive videos is prohibitively expensive.
Therefore, with only limited annotated data, learning discriminative features via deep neural network
can lead to severe overfitting and slow convergence. To tackle these issues, previous works have introduced effective
practical techniques such as ReLU \cite{relu} and Drop-out \cite{dropout} to improve the performance of neural networks, but
have not considered directly improving the
discriminative capability of neurons. The features from a CNN are learned by back-propagating prediction error from the output layer
\cite{bp}, and hidden layers receive no direct guidance on class information.
Worse, in very deep networks, the early hidden layers often suffer from vanishing gradients, which leads to slow optimization convergence
and the network converging to a poor local minimum. Therefore, the quality of the learned features of the hidden layers
might be potentially diminished~\cite{fergus14,vanish}.

To tackle these problems, we propose a new supervised deep neural network, \textit{Label Consistent Neural Network}, to
learn discriminative features for recognition.
Our approach provides explicit supervision, \textit{i.e.} label information, to late hidden
layers, by incorporating a label consistency constraint called ``discriminative representation error'' loss, which is
combined with the classification loss to form the overall objective function. The benefits of our approach are two-fold:
(1) with explicit supervision to hidden layers, the problem of vanishing gradients can be alleviated and faster convergence
is observed; (2) more discriminative late hidden layer features lead to increased discriminative power of classifiers at the output layer;
interestingly, the learned discriminative features alone can achieve good classification
performance even with a simple $k$-NN classifier. In practice, our new formulation can be easily incorporated into
any neural network trained using backpropagation. Our approach is evaluated on publicly available action and object recognition
datasets. Although we only present experimental results for action and object recognition, the method can be applied to
other tasks such as image retrieval, compression, restorations \textit{etc}., since it
generates class-specific compact representations.

\subsection{Main Contributions} \label{sec1_1}

The main contributions of LCNN are three-fold.
\begin{itemize}
\item By adding explicit supervision to late hidden layers via a ``discriminative representation
error'', LCNN learns more discriminative
features resulting in better classifier training at the output layer. The representations generated
by late hidden layers are discriminative enough to achieve good performance using a simple $k$-NN classifier.

\item The label consistency constraint alleviates the problem of vanishing gradients and leads to faster
convergence during training, especially when limited training data is available.

\item We achieve state-of-the-art performance on several action and object category recognition tasks, and the compact
class-specific representations generated by LCNN can be directly used in other applications.
\end{itemize}

\section{Related Work}\label{sec2}

CNNs have achieved performance improvements over traditional
hand-crafted features in image recognition \cite{hinton12}, detection \cite{rcnn} and retrieval
\cite{cnnretrieval} \textit{etc}. This is due to the availability of large-scale image
datasets \cite{imagenet} and recent technical improvements such as ReLU \cite{relu}, drop-out \cite{dropout}, $1\times 1$
convolution \cite{nin,googlenet}, batch normalization \cite{batchnorm} and
data augmentation based on random flipping, RGB jittering, contrast normalization \cite{hinton12,nin}, which
helps speed up convergence while avoiding overfitting.

AlexNet~\cite{hinton12} initiated the dramatic performance improvements of CNN in static image recognition and current
state-of-the-art performance has been obtained by deeper and more sophisticated network architectures such as VGGNet \cite{vgg} and GoogLeNet \cite{googlenet}.
Very recently, researchers have applied CNNs to action and event recognition in videos. While initial approaches use
image-trained CNN models to extract frame-level features and aggregate them into video-level
descriptors \cite{joe15,shengxin15,cnnfisher}, more recent work trains CNNs using video data and focuses on effectively incorporating the
temporal dimension and learning good spatial-temporal features automatically
\cite{3dcnn,karpathy14,twostream1,twostream2,twostream3,trajpool}. Two-stream CNNs \cite{twostream1} are perhaps the
most successful architecture for action recognition currently. They consist of a
spatial net trained with video frames and a temporal net trained with optical flow fields. With the two streams capturing
spatial and temporal information separately, the late fusion of the two produces competitive action recognition
results. \cite{twostream2} and \cite{twostream3} have obtained further performance gain by exploring deeper two-stream network architectures and
refining technical details; \cite{trajpool} achieved state-of-the-art in action recognition by integrating two-stream CNNs, improved trajectories and Fisher Vector encoding.

It is also worth comparing our LCNN with limited prior work which aims to improve the discriminativeness of learned features. \cite{pretrain}
performs greedy layer-wise supervised pre-training as initialization and fine-tunes the parameters of all layers together. Our work
introduces the supervision to intermediate layers as part of the objective function during training and can be optimized by
backpropagation in an integrated way, rather than layer-wise greedy pretraining and then fine-tuning. \cite{targetcoding}
replaces the output softmax layer with an error-correcting coding layer to produce error correcting codes as network
output. Their network is still trained by back-propagating the error at the output and no direct supervision
is added to hidden layers. Deeply Supervised Net (DSN) \cite{dsn} introduces an SVM classifier for \textit{each} hidden layer, and the final objective
function is the linear combination of the prediction losses at all hidden layers and output layer. Using all-layer supervision, balancing between multiple losses might be challenging and the network is non-trivial to tune, since only the classifier at the output layer will be used at test time and the effects of the classifiers at hidden layers are difficult to evaluate. Similarly, \cite{Sun15} also adds identification and verification supervisory signals to each hidden layer to extract face representations. In our work, instead of adding a prediction loss to each hidden layer, we introduce a novel representation loss to guide the format of the learned features at late hidden layers only, since early layers of CNNs tend to capture low-level edges, corners and mid-level parts
and they should be shared across categories, while the late hidden layers are more class-specific \cite{fergus14}.

\begin{figure*}
\begin{center}
\includegraphics[width=0.7\linewidth]{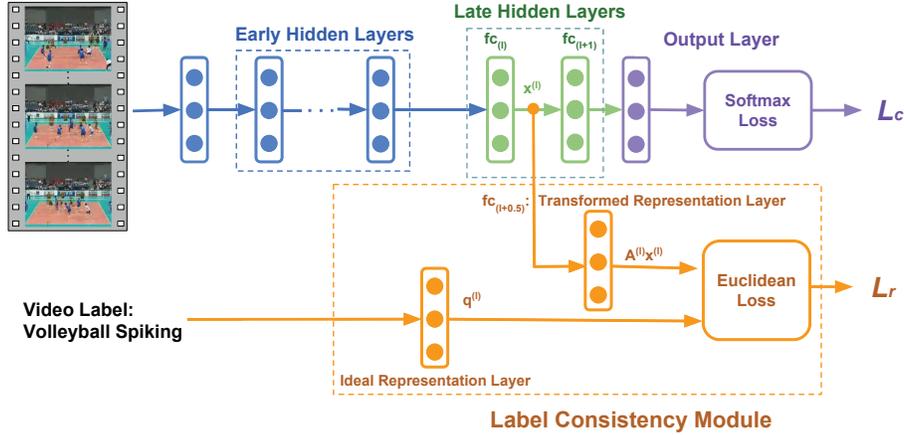}
\end{center}
\caption{An example of the LCNN structure. The label consistency module is added to the $l^{\text{th}}$ hidden layer,
which is a fully-connected layer $\text{fc}_l$. Its representation $\mathbf{x}^l$ is transformed to be
$\mathbf{A}^{(l)}$$\mathbf{x}^l$, which is the output of the transformed representation layer $\text{fc}_{l+0.5}$. Note that the applicability of the proposed label consistency module is not limited to fully-connected layers.}
\label{fig1}
\vspace{-0.5cm}
\end{figure*}

\section{Feature Learning via Supervised Deep Neural Network} \label{sec3}

Let $(\mathbf{x}, y)$ denote a training sample $\mathbf{x}$ and its label $y$. For a CNN with $n$ layers, let
$\mathbf{x}^{(i)}$ denote the output of the $i^{\text{th}}$ layer and $L_c$ its objective function.
$\mathbf{x}^{(0)} = \mathbf{x}$ is the input data and $\mathbf{x}^{(n)}$ is the output of the network.
Therefore, the network architecture can be concisely expressed as
\begin{align}
&\mathbf{x}^{(i)} = F(\mathbf{W}^{(i)}\mathbf{x}^{(i - 1)}),\quad i = 1, 2, ..., n \label{eq1} \\
&L_c = L_c(\mathbf{x}, y, \mathbf{W}) = C(\mathbf{x}^{(n)}, y), \label{eq2}
\end{align}
where $\mathbf{W}^{(i)}$ represents the network parameters of the $i^{\text{th}}$ layer, $\mathbf{W}^{(i)}\mathbf{x}^{(i - 1)}$
is the linear operation (\textit{e.g.} convolution in convolutional layer, or linear transformation in fully-connected layer), and
$\mathbf{W} = \{\mathbf{W}^{(i)}\}_{i = 1, 2, \ldots, n}$;
 $F(\cdotp)$ is a non-linear activation function (\textit{e.g.} ReLU); $C(\cdotp)$ is a prediction error such as softmax
loss.
The network is trained with back-propagation, and the
gradients are computed as:
\begin{align}
\frac {\partial L_c} {\partial \mathbf{x}^{(i)}} &=
\begin{cases}
\frac {\partial C(\mathbf{x}^{(n)}, y)} {\partial \mathbf{x}^{(n)}},\quad i = n\\
\frac {\partial L_c} {\partial \mathbf{x}^{(i + 1)}} \frac {\partial F(\mathbf{W}^{(i + 1)}\mathbf{x}^{(i)})} {\partial
\mathbf{x}^{(i)}}, i \neq n\label{eq4}
\end{cases} \\
\frac {\partial L_c} {\partial \mathbf{W}^{(i)}} &= \frac {\partial L_c} {\partial \mathbf{x}^{(i)}} \frac {\partial
F(\mathbf{W}^{(i)}\mathbf{x}^{(i - 1)})} {\partial \mathbf{W}^{(i)}}, \label{eq5}
\end{align}
where $i = 1, 2, 3, ..., n$.

\section{Label Consistent Neural Network (LCNN)} \label{sec4}

\subsection{Motivation} \label{sec4_1}
The sparse representation for classification assumes that a testing sample can be well represented by training samples
from the same class \cite{yima}. Similarly, dictionary learning for recognition maintains label
information for dictionary items during training in order to generate discriminative or class-specific sparse codes
\cite{lcksvd,fisherdl}. In a neural network, the representation of a certain layer is
generated by the neuron activations in that layer. If the class distribution for each neuron is highly peaked in one class, it
enforces a label consistency constraint on each neuron. This leads to a discriminative representation over learned
class-specific neurons.

It has been observed that early hidden layers of a CNN tend to capture low-level features shared across categories such as edges
and corners, while late hidden layers are more class-specific \cite{fergus14}. To improve the discriminativeness of features, LCNN adds explicit
supervision to late hidden layers; more specifically, we associate each neuron to a certain class label and ideally the
neuron will only activate when a sample of the corresponding class is presented. The label consistency constraint on neurons
in LCNN will be imposed by introducing a ``discriminative representation error'' loss on late hidden layers, which
will form part of the objective function during training.

\begin{figure*}[t]
\begin{center}
\hspace{-0.75 cm}
    \label{fig2:j1}
    \includegraphics[width=0.033\linewidth]{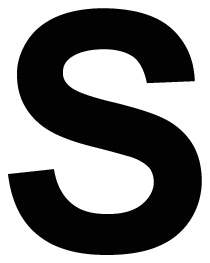}
\hspace{-0.22 cm}
    \label{fig2:a1}
    \includegraphics[width=0.13\linewidth]{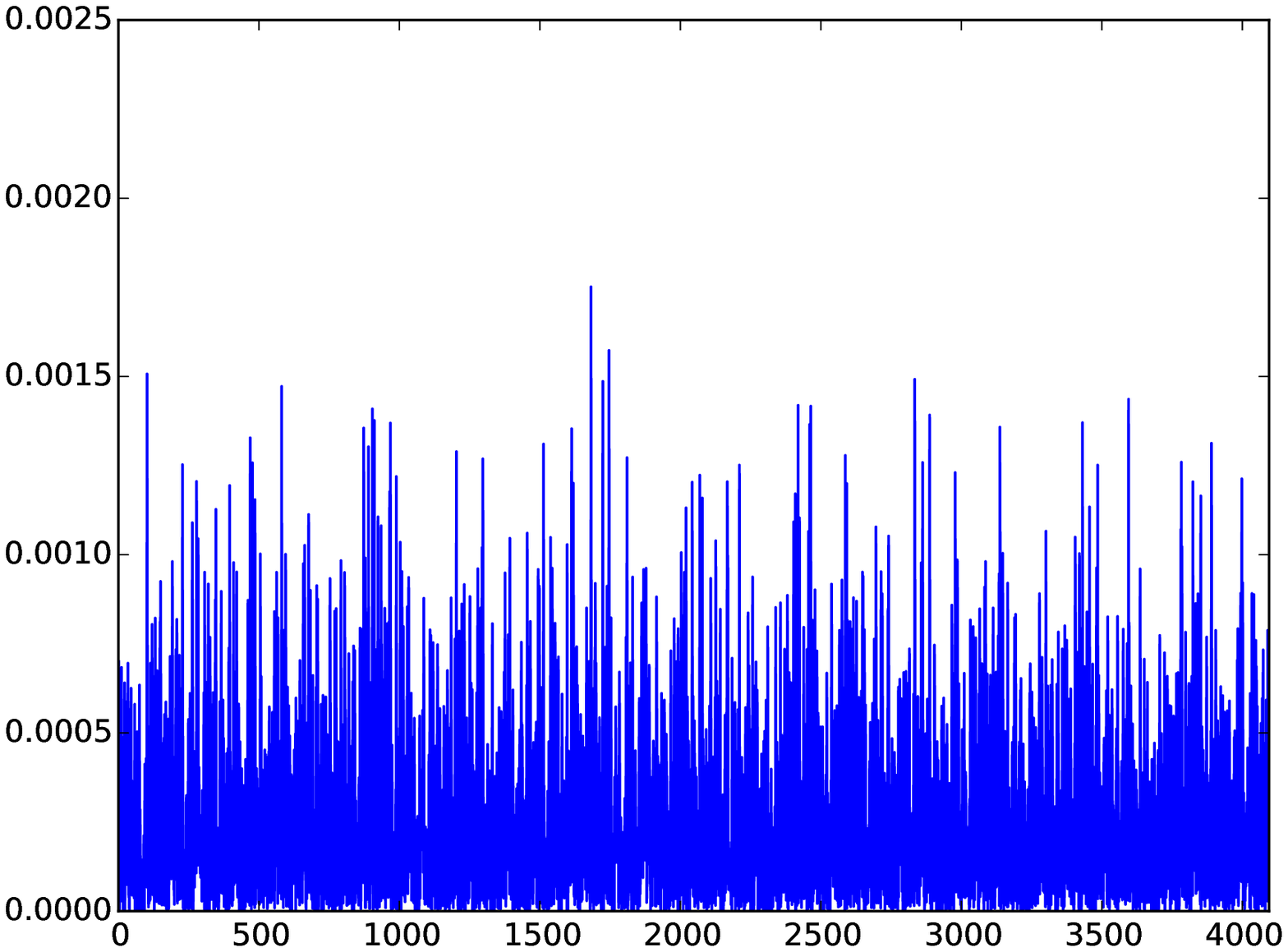}
\hspace{-0.29cm}
    \label{fig2:b1}
    \includegraphics[width=0.1\linewidth]{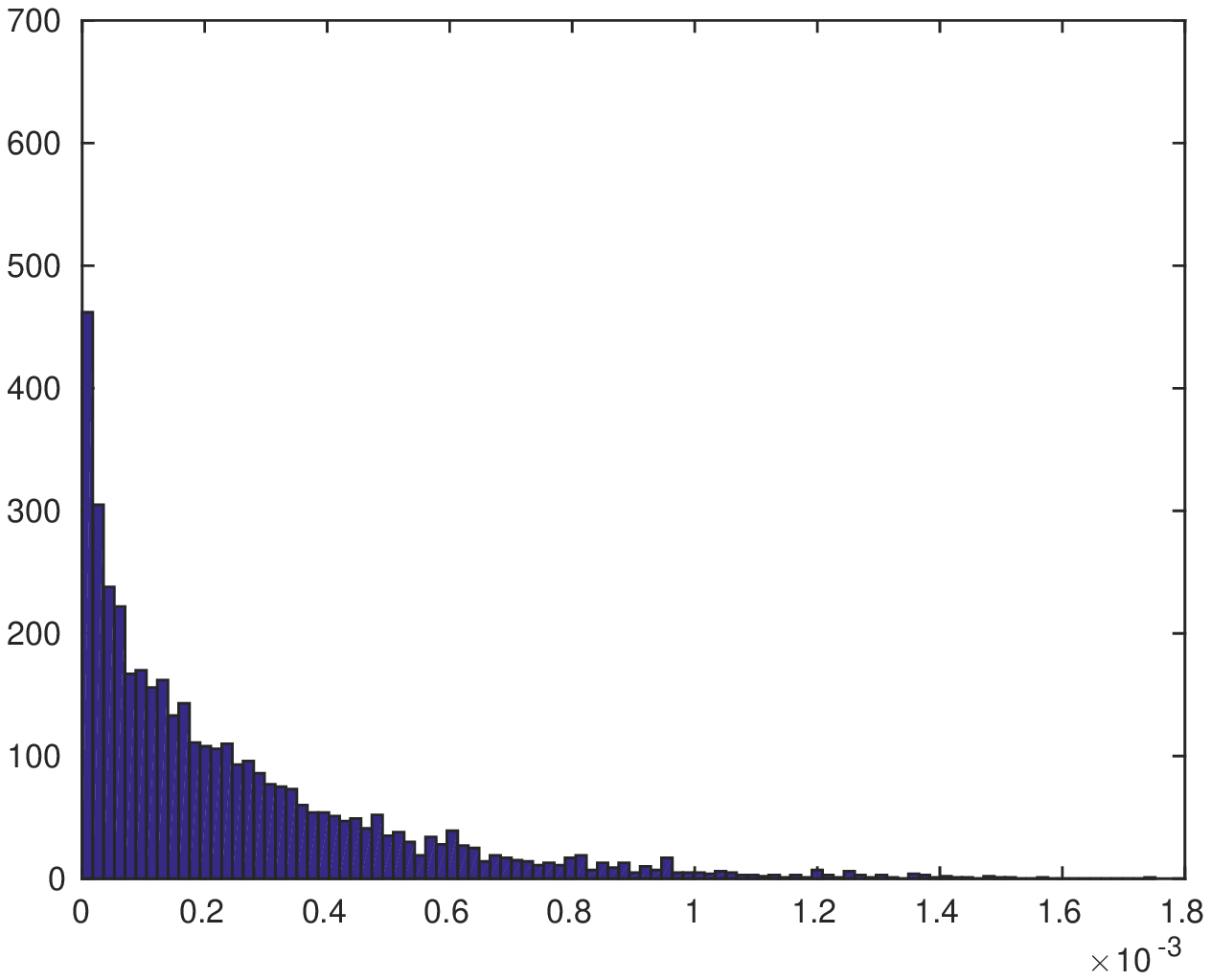}
\hspace{-0.26cm}
    \label{fig2:c1}
    \includegraphics[width=0.13\linewidth]{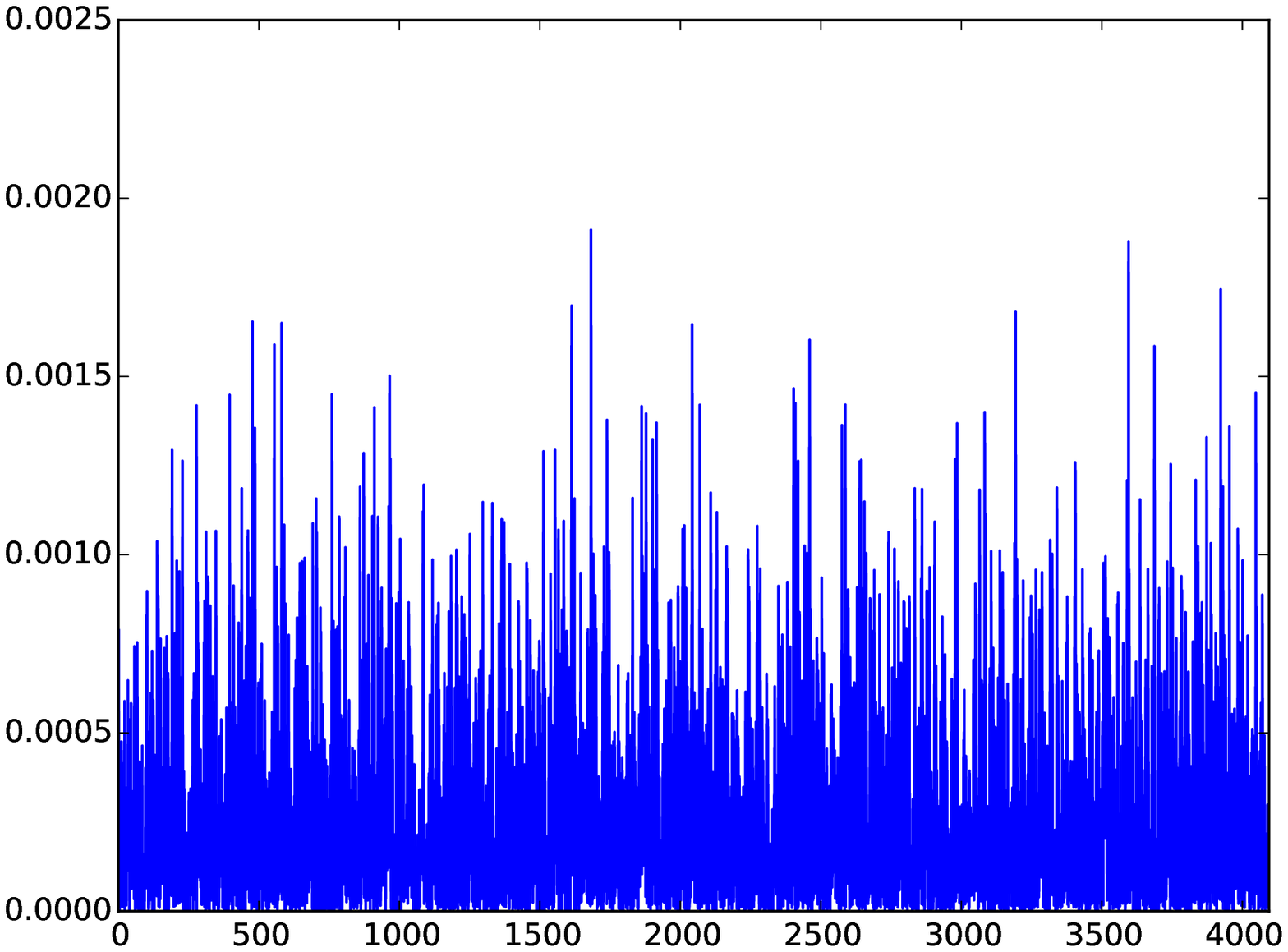}
\hspace{-0.29cm}
    \label{fig2:d1}
    \includegraphics[width=0.1\linewidth]{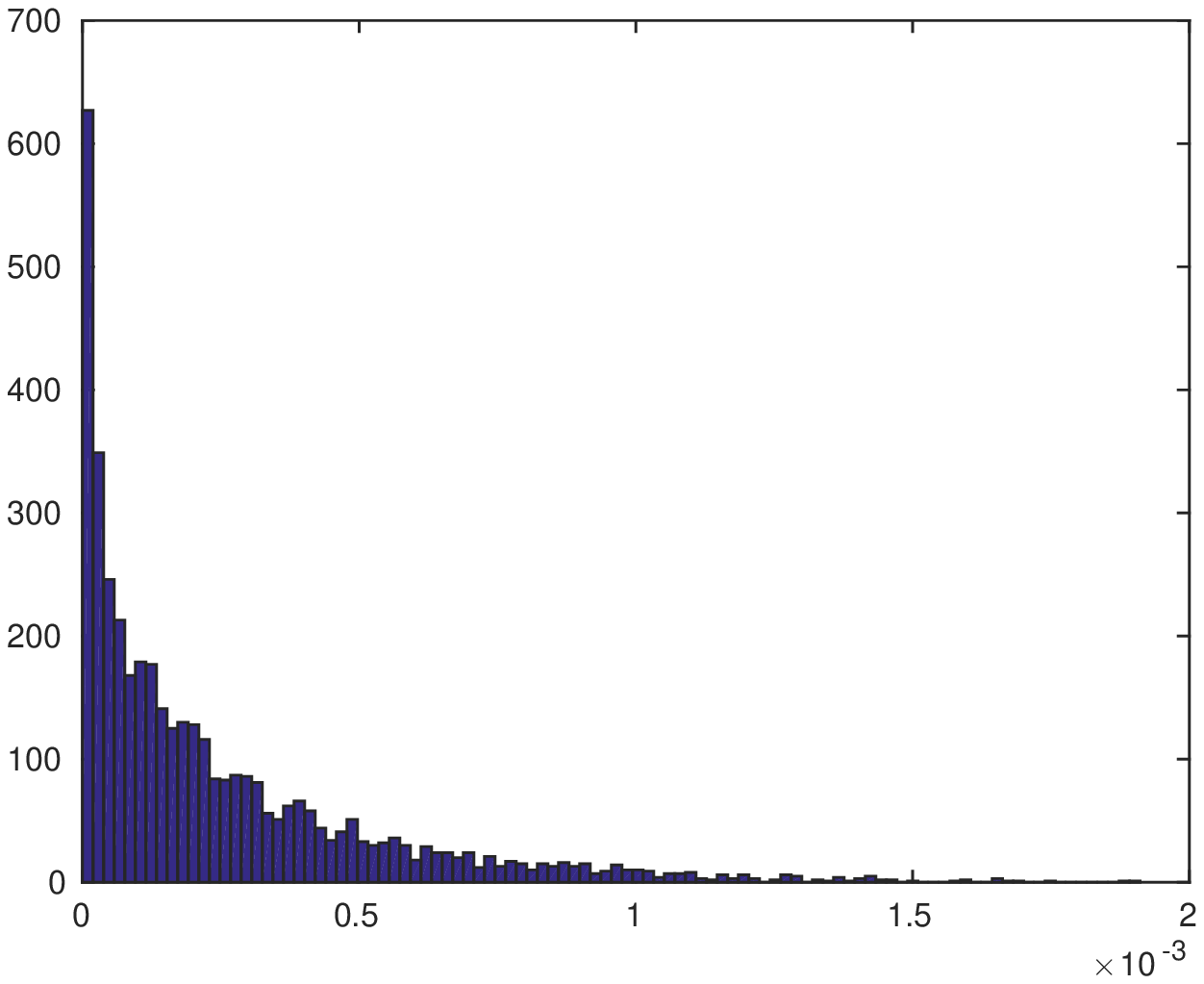}
\hspace{-0.26cm}
    \label{fig2:e1}
    \includegraphics[width=0.13\linewidth]{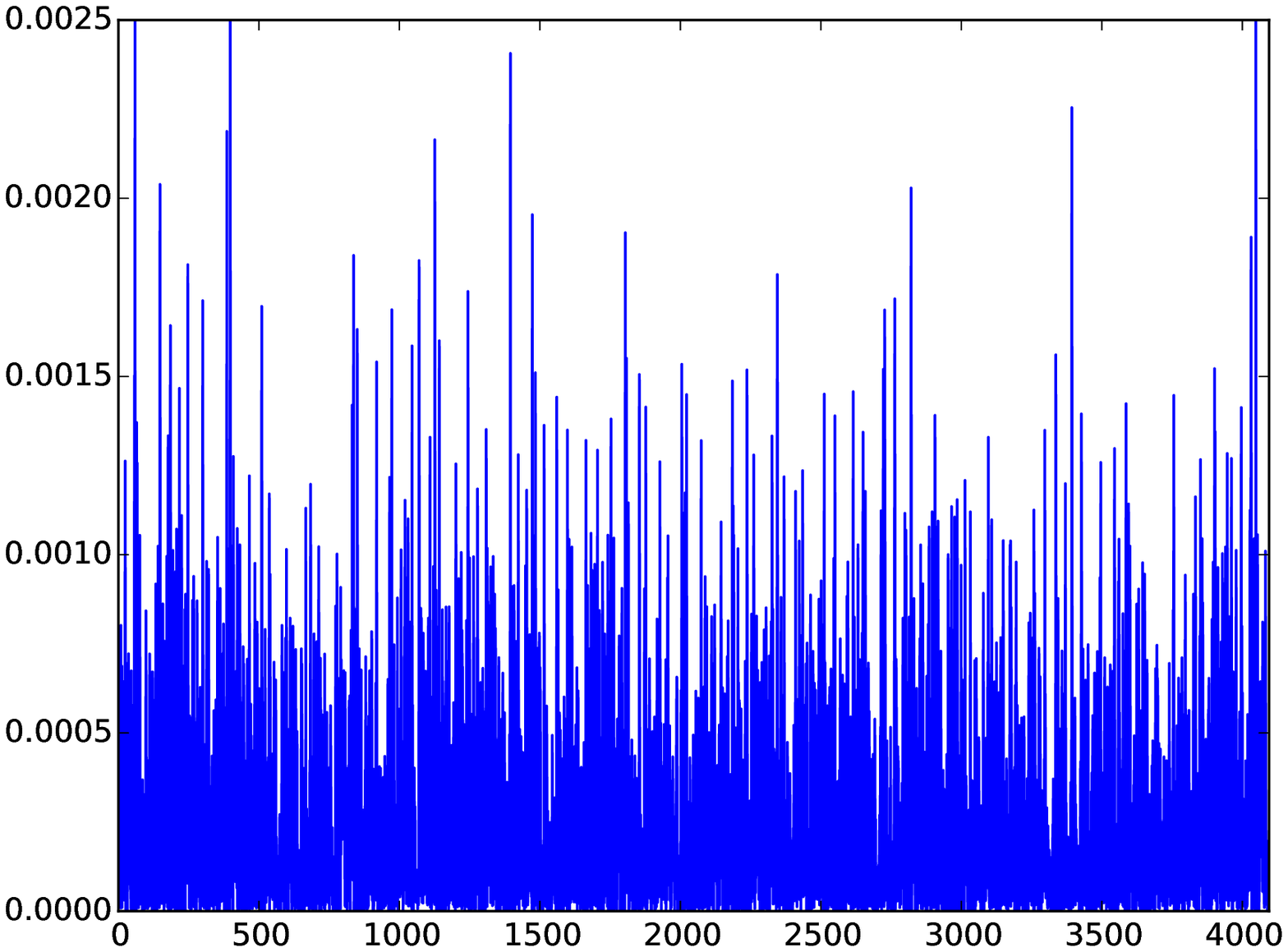}
\hspace{-0.29cm}
    \label{fig2:f1}
    \includegraphics[width=0.1\linewidth]{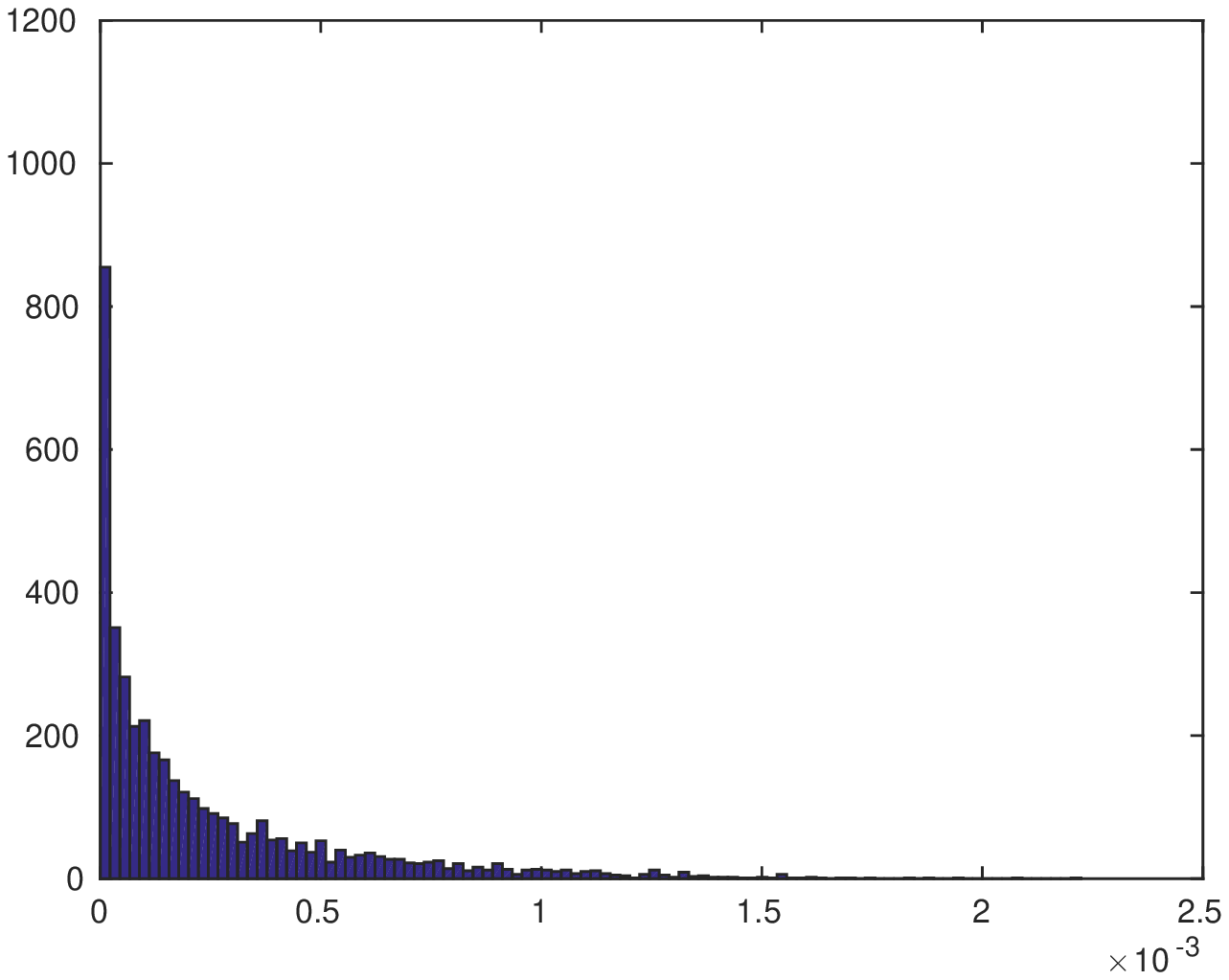}
\hspace{-0.26cm}
    \label{fig2:g1}
    \includegraphics[width=0.13\linewidth]{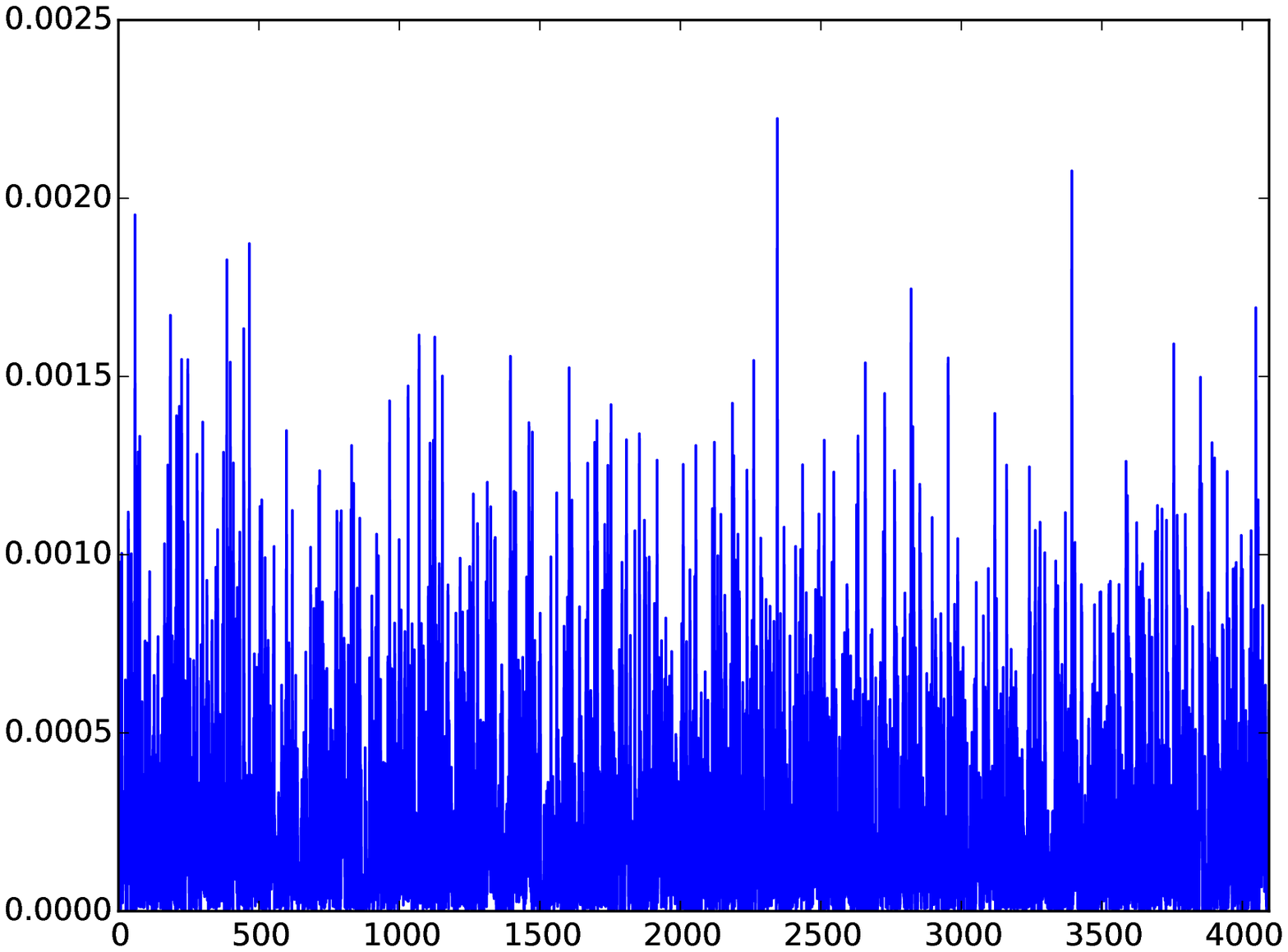}
\hspace{-0.29cm}
    \label{fig2:h1}
    \includegraphics[width=0.1\linewidth]{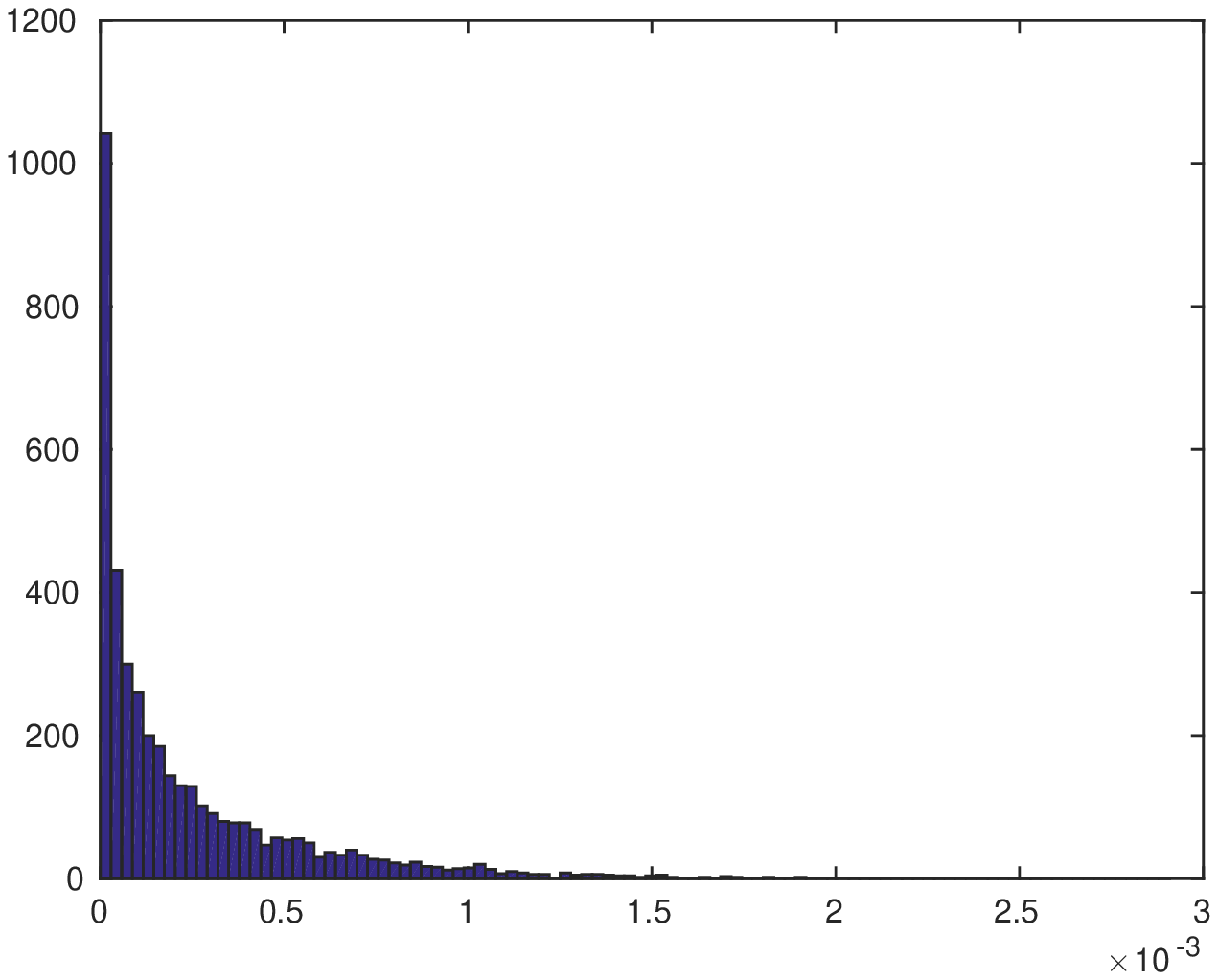}
\hspace{-0.26cm}
    \label{fig2:i1}
    \includegraphics[width=0.13\linewidth]{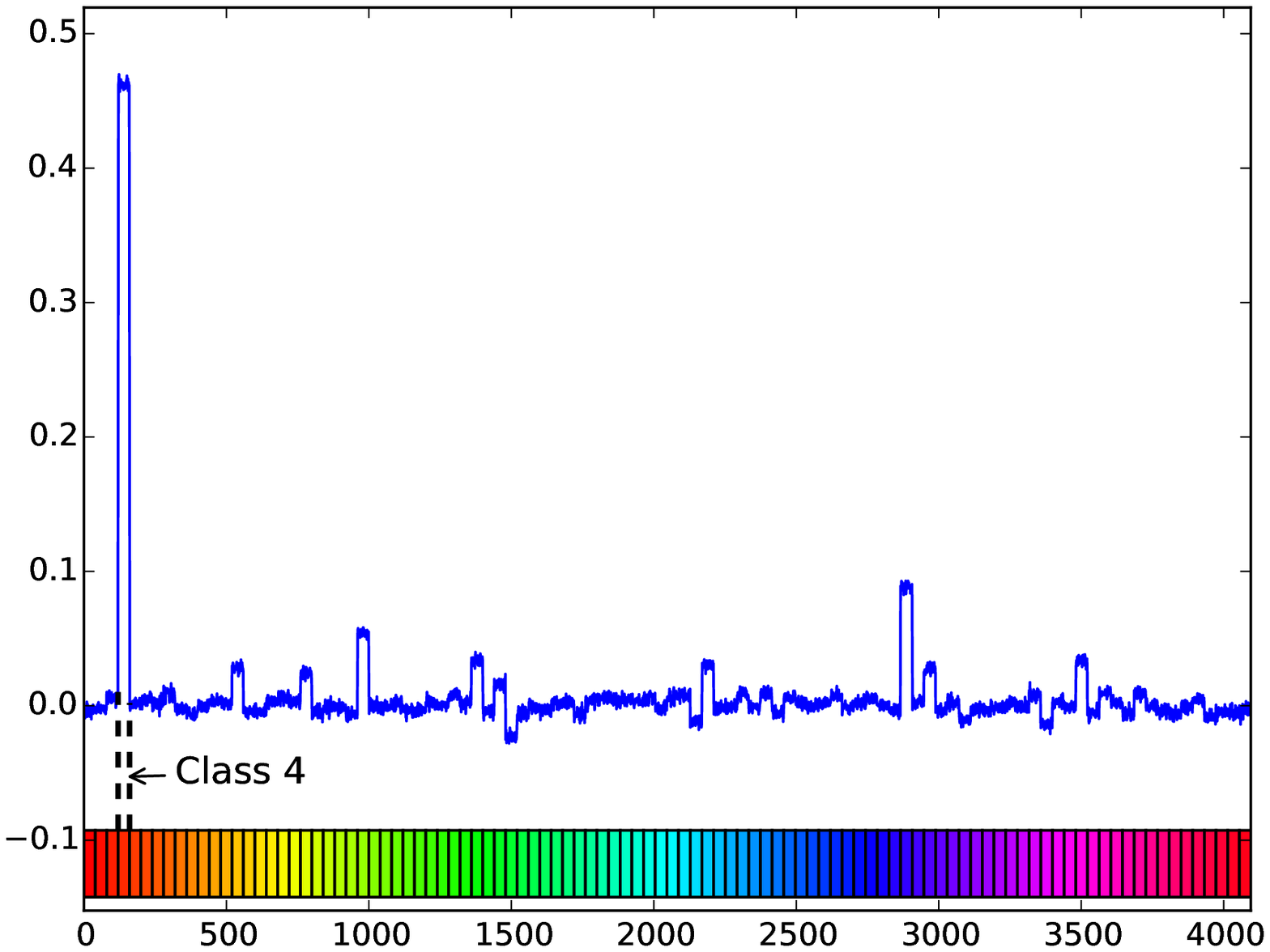}
\\
\hspace{-0.75cm}
    \label{fig2:j2}
    \includegraphics[width=0.033\linewidth]{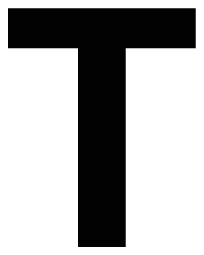}
\hspace{-0.22cm}
    \label{fig2:a2}
    \includegraphics[width=0.13\linewidth]{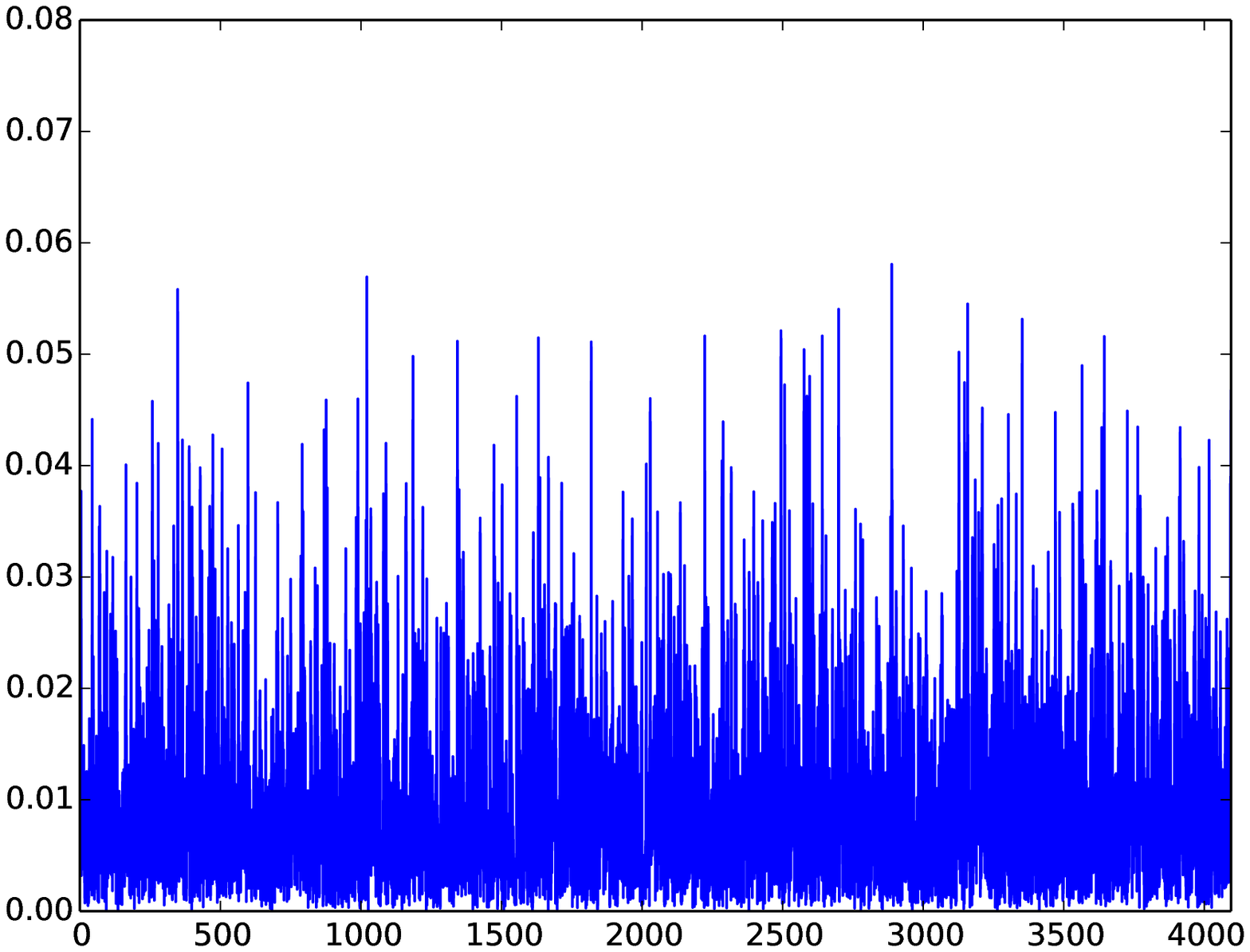}
\hspace{-0.29cm}
    \label{fig2:b2}
    \includegraphics[width=0.1\linewidth]{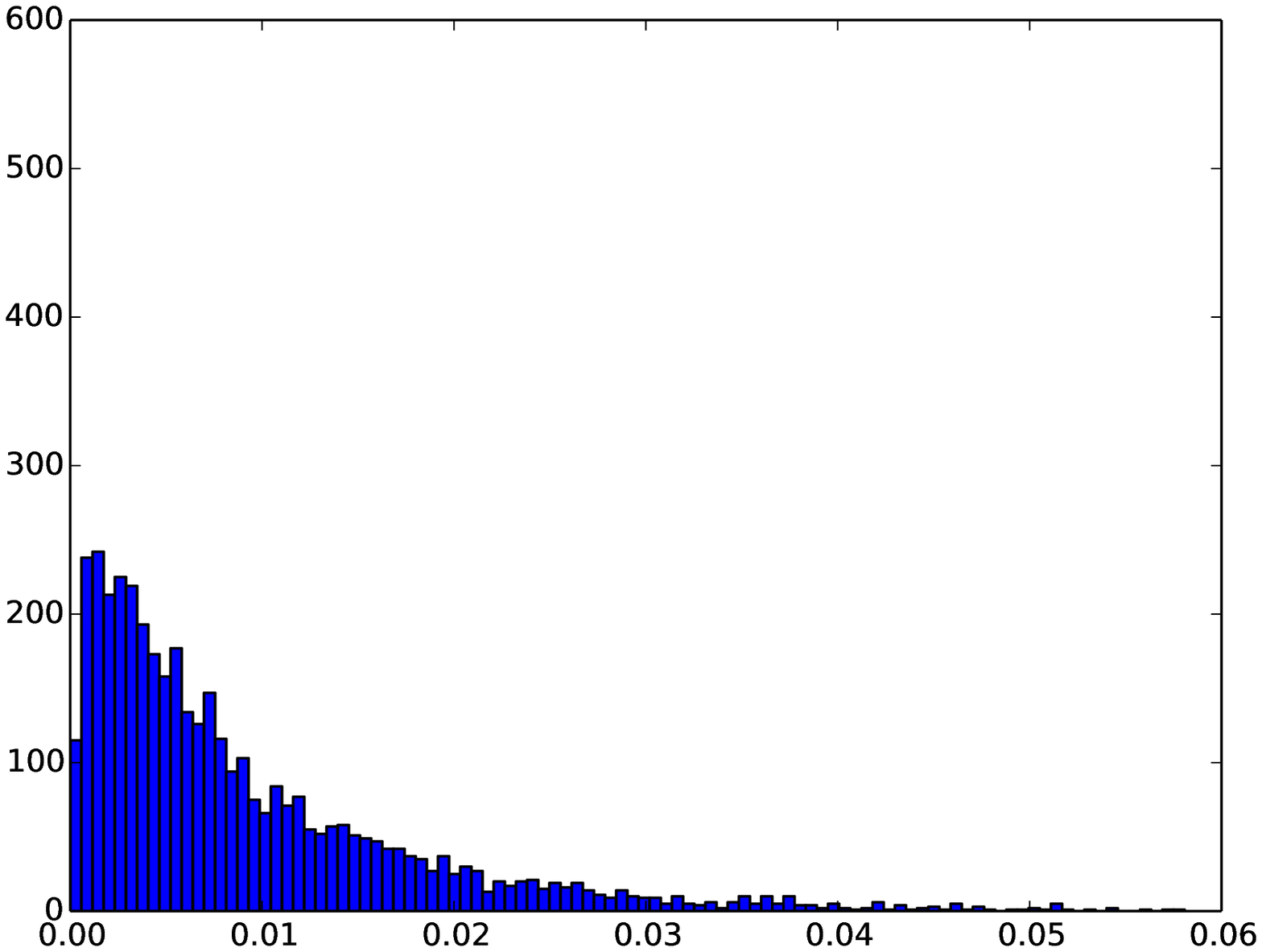}
\hspace{-0.26cm}
    \label{fig2:c2}
    \includegraphics[width=0.13\linewidth]{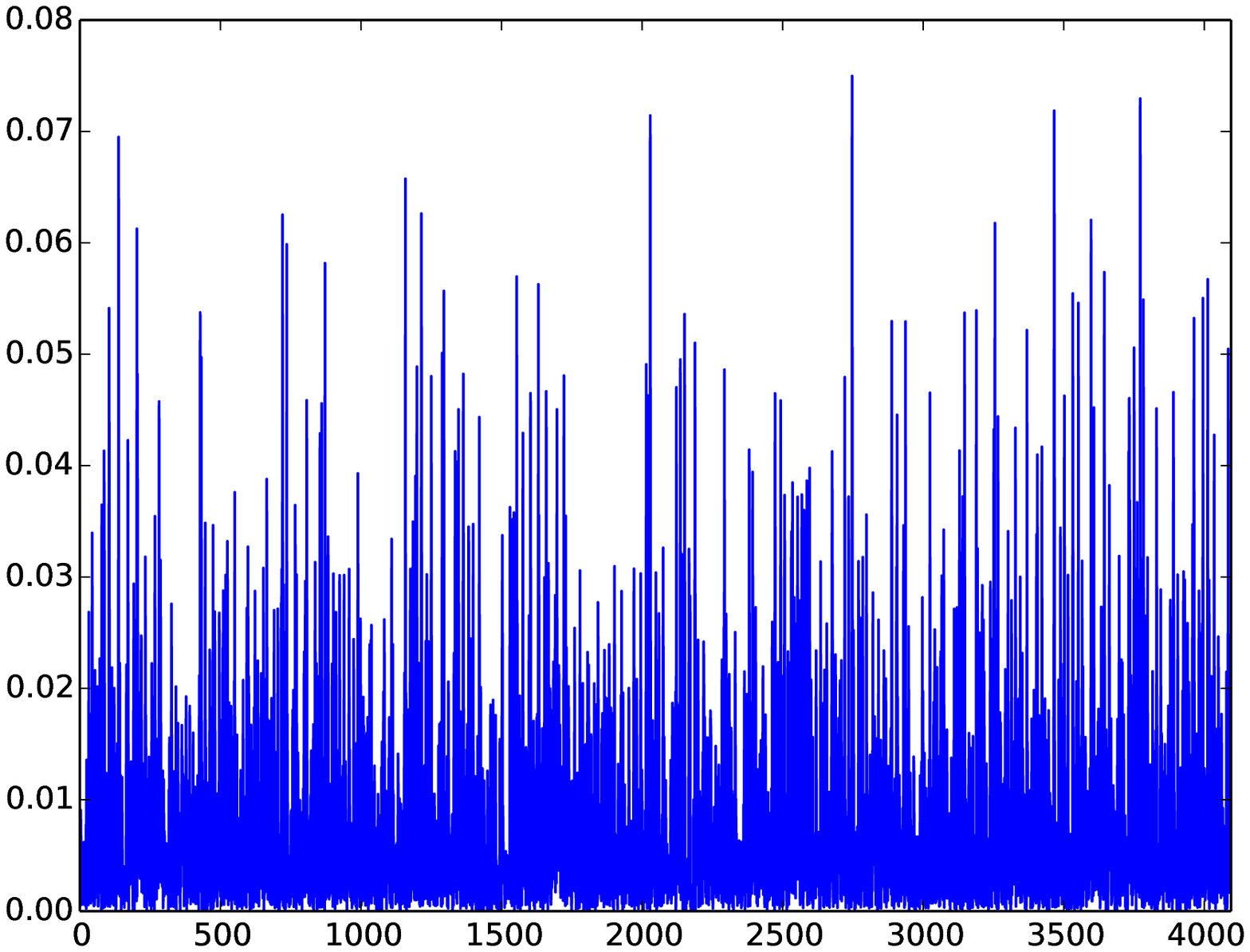}
\hspace{-0.29cm}
    \label{fig2:d2}
    \includegraphics[width=0.1\linewidth]{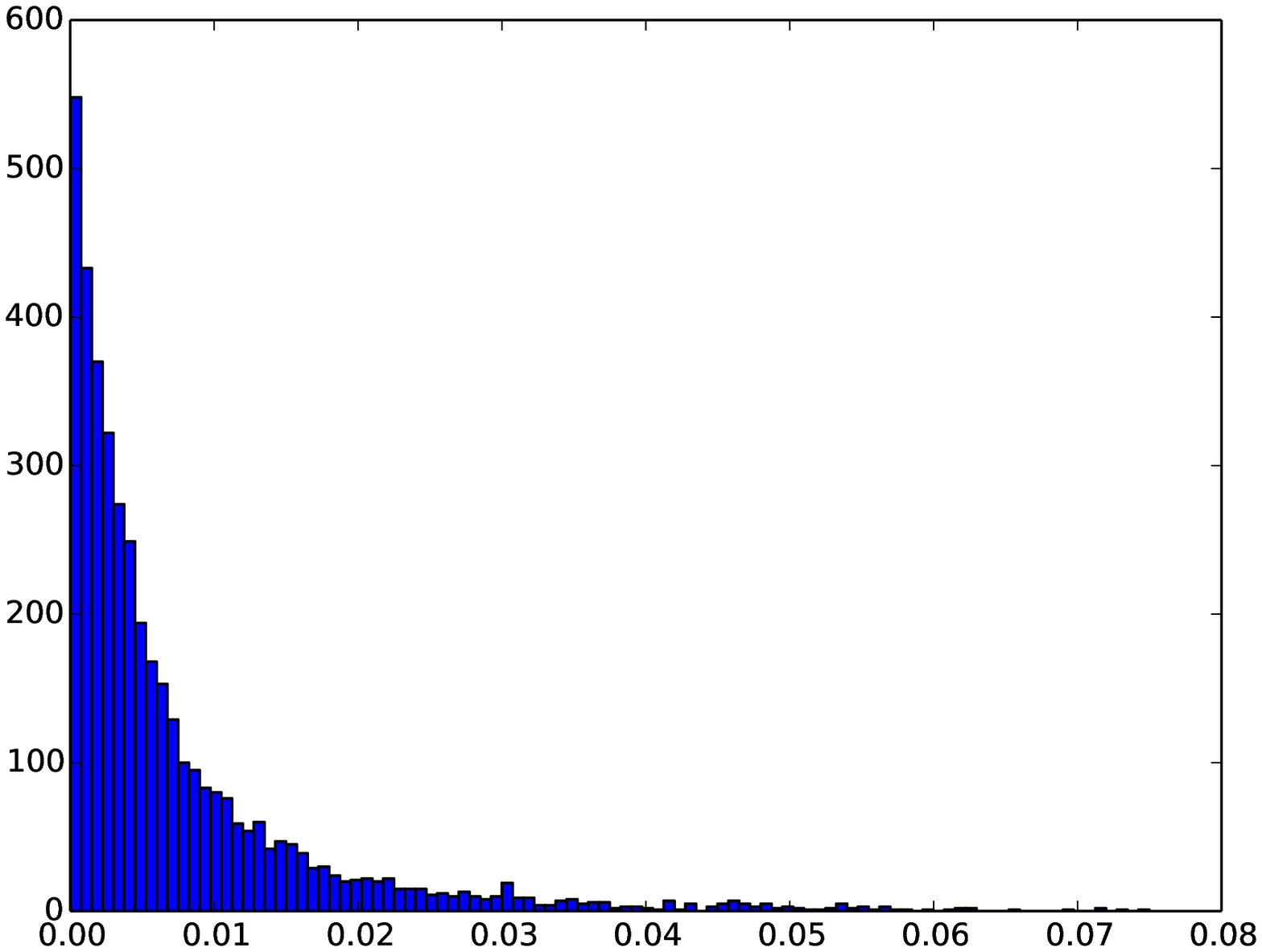}
\hspace{-0.26cm}
    \label{fig2:e2}
    \includegraphics[width=0.13\linewidth]{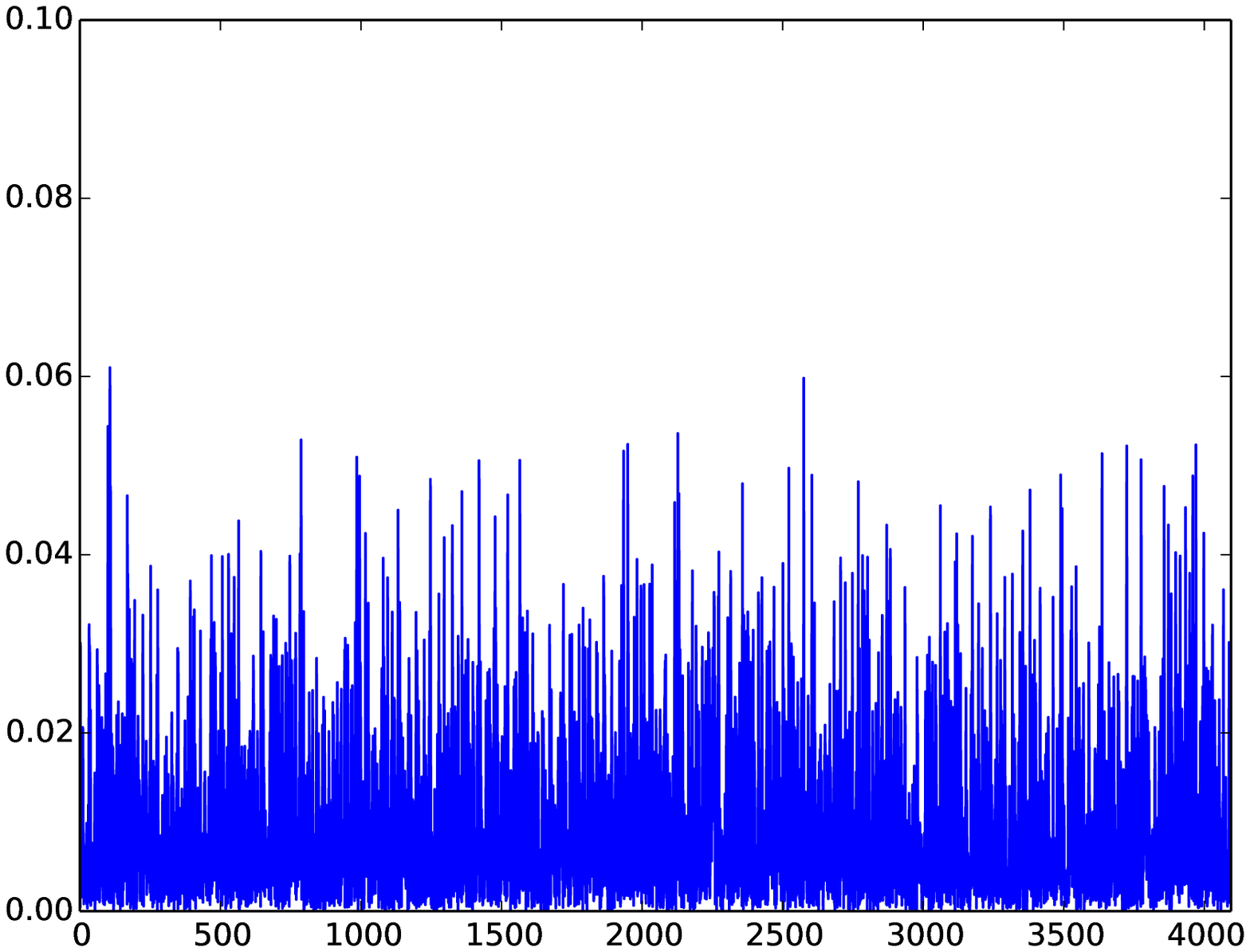}
\hspace{-0.29cm}
    \label{fig2:f2}
    \includegraphics[width=0.1\linewidth]{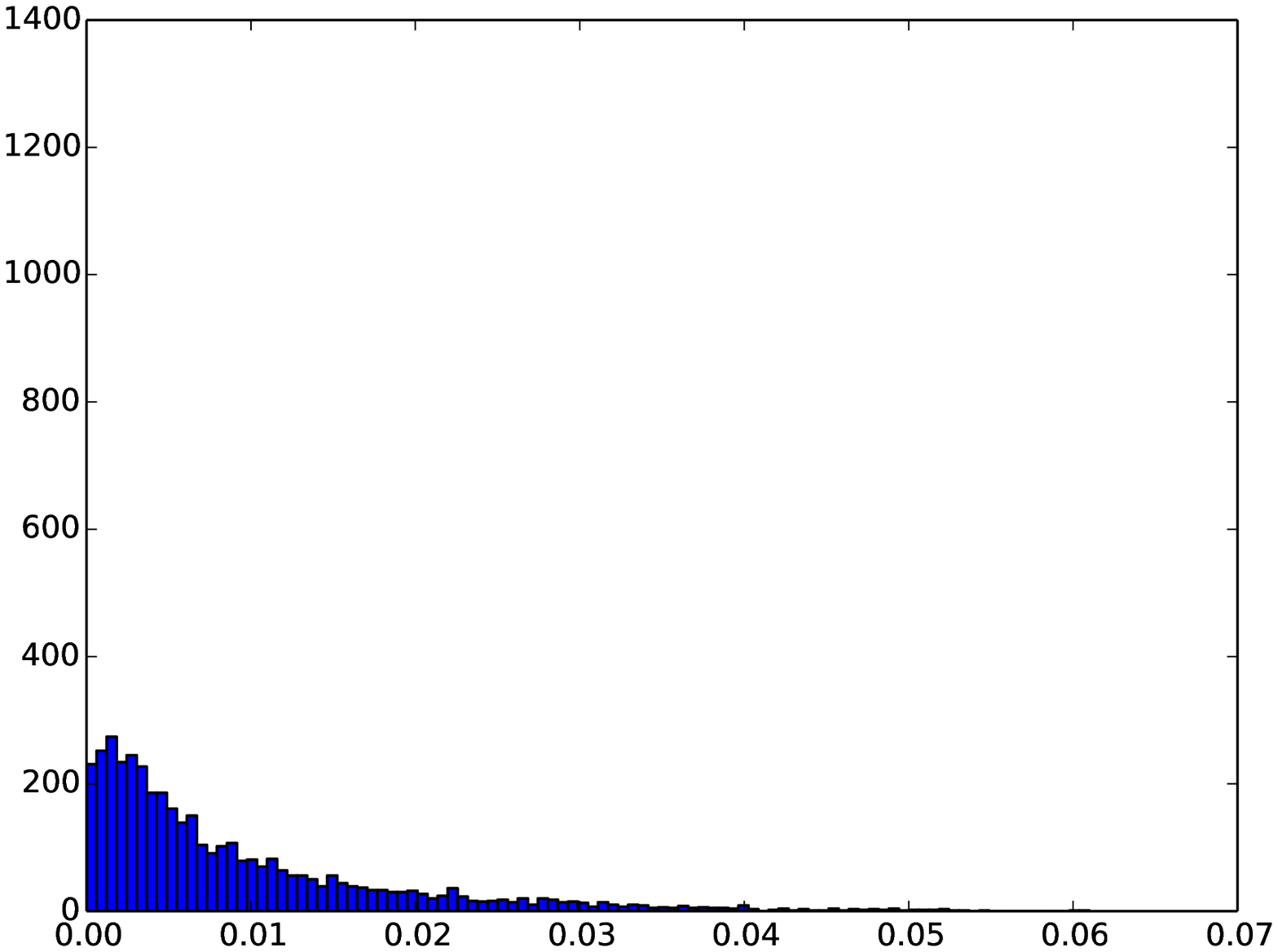}
\hspace{-0.26cm}
    \label{fig2:g2}
    \includegraphics[width=0.13\linewidth]{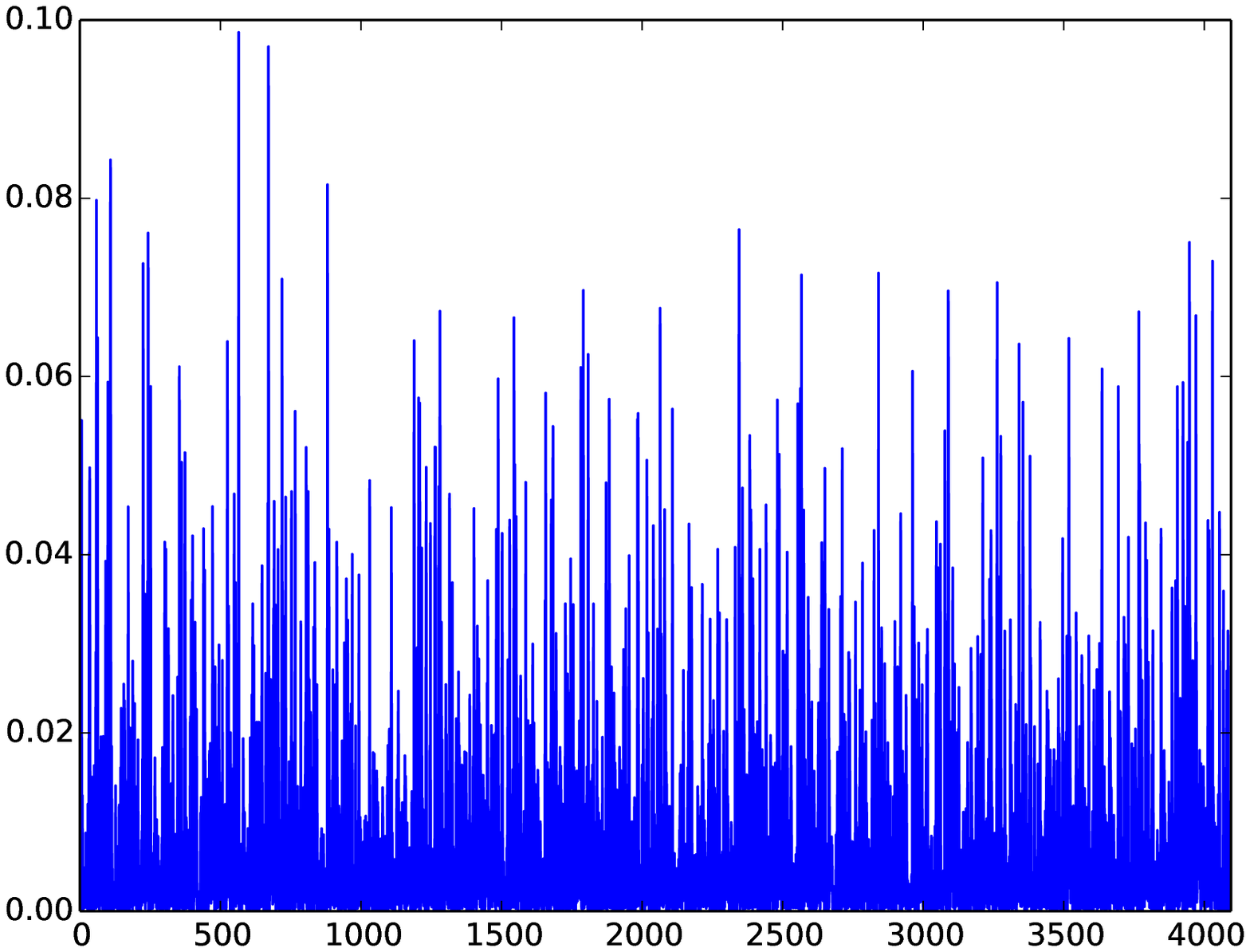}
\hspace{-0.29cm}
    \label{fig2:h2}
    \includegraphics[width=0.1\linewidth]{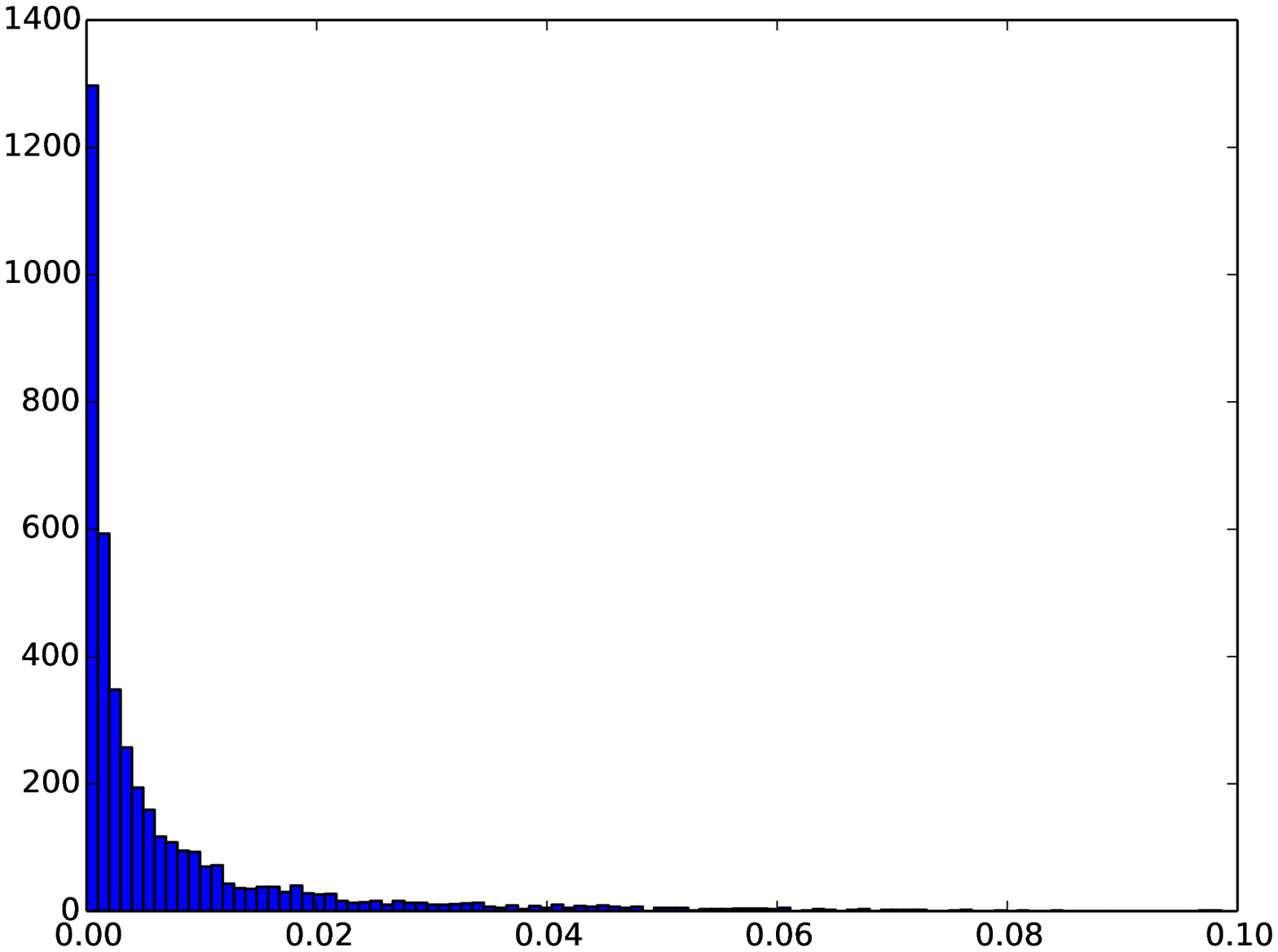}
\hspace{-0.26cm}
    \label{fig2:i2}
    \includegraphics[width=0.13\linewidth]{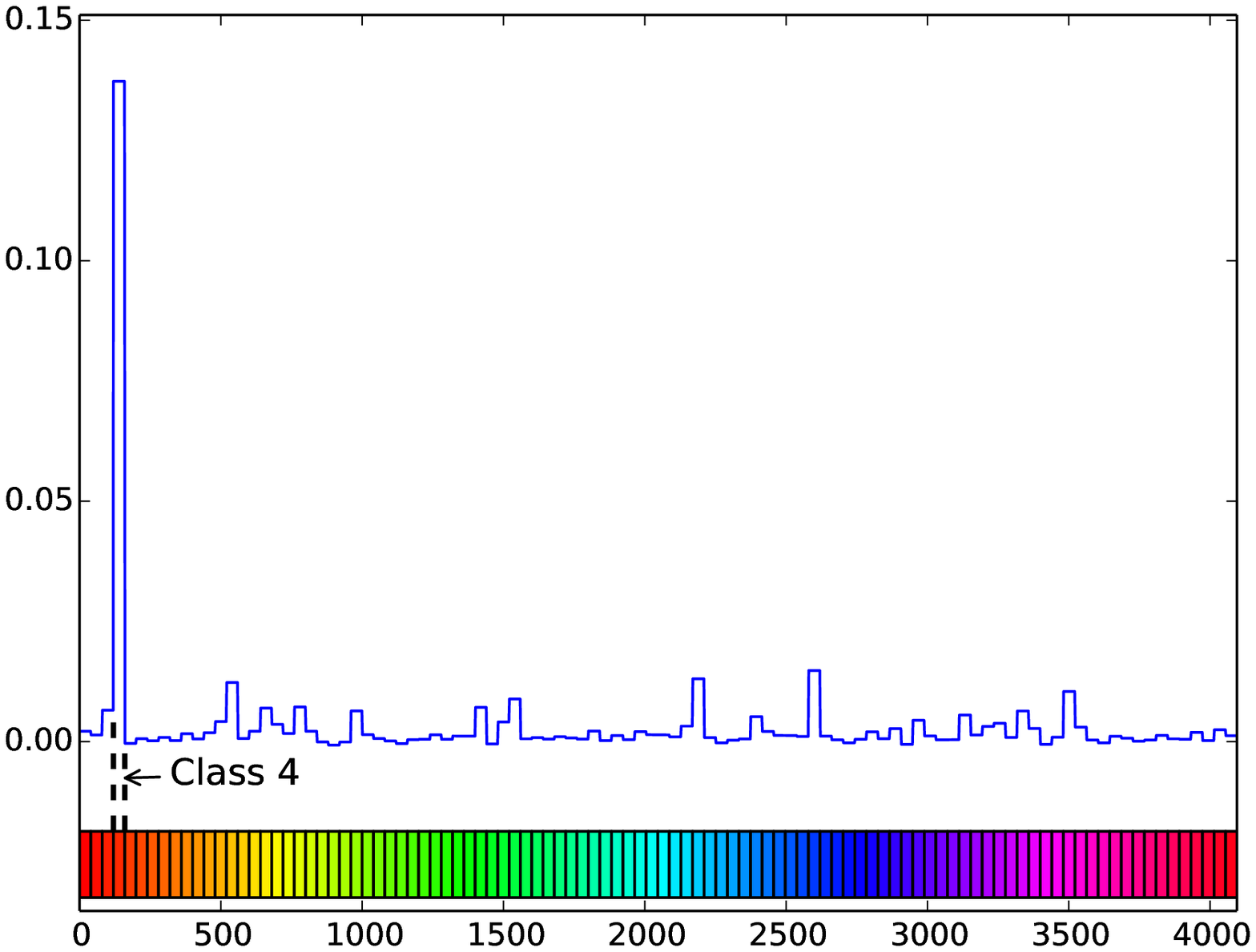}
\\
\hspace{-0.75cm}
    \label{fig2:j3}
    \includegraphics[width=0.033\linewidth]{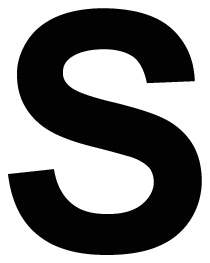}
\hspace{-0.22cm}
    \label{fig2:a3}
    \includegraphics[width=0.13\linewidth]{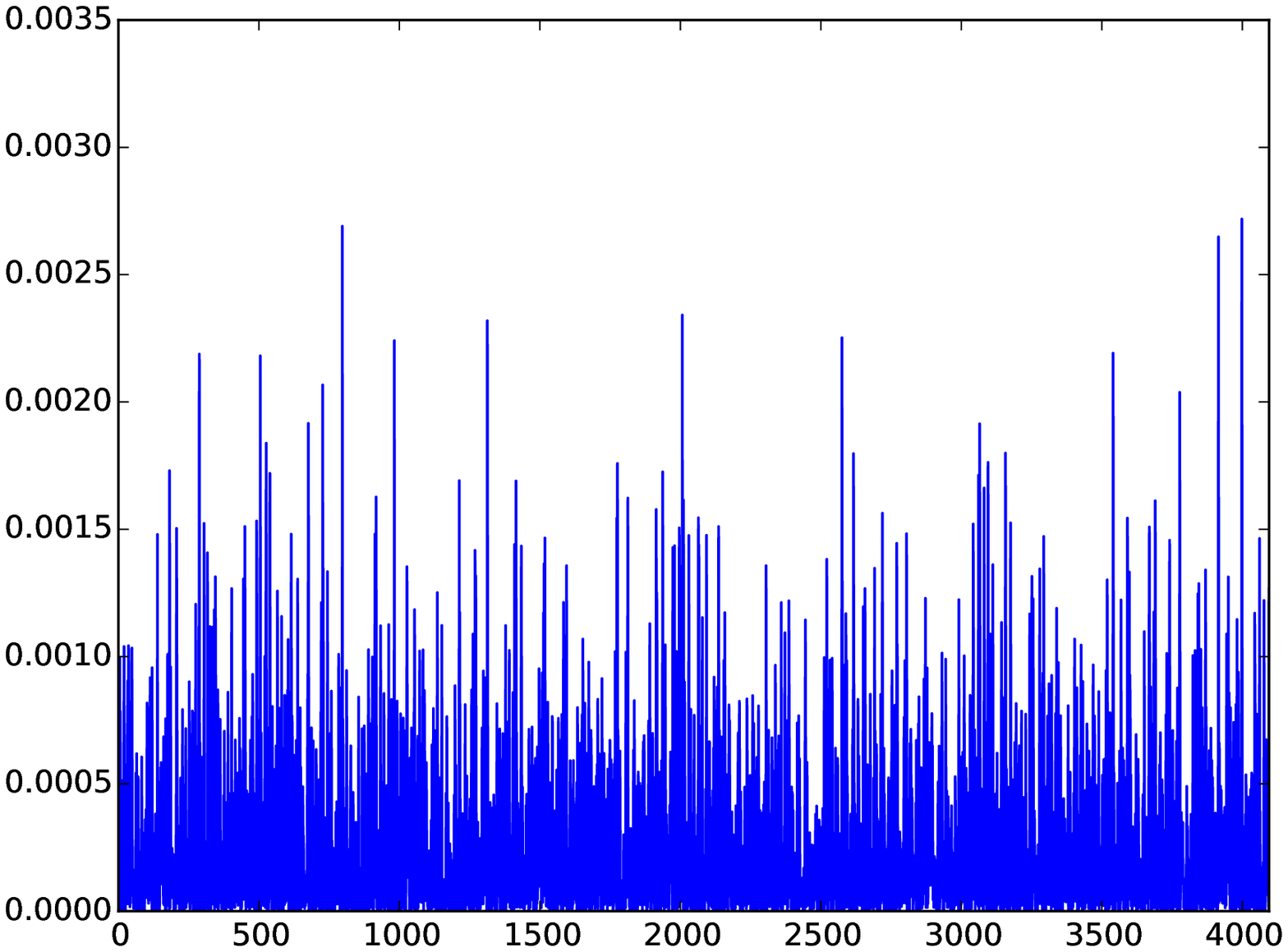}
\hspace{-0.29cm}
    \label{fig2:b3}
    \includegraphics[width=0.1\linewidth]{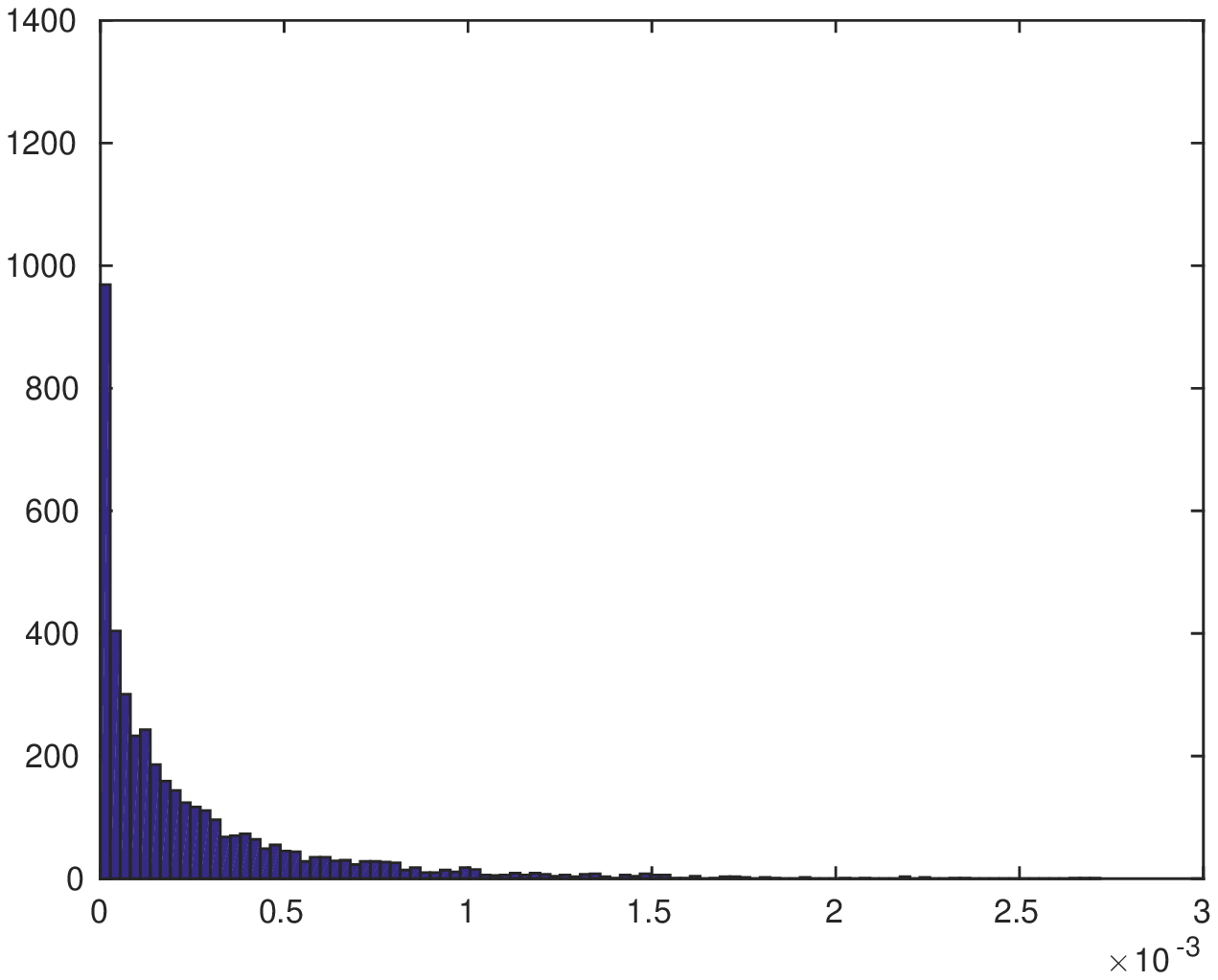}
\hspace{-0.26cm}
    \label{fig2:c3}
    \includegraphics[width=0.13\linewidth]{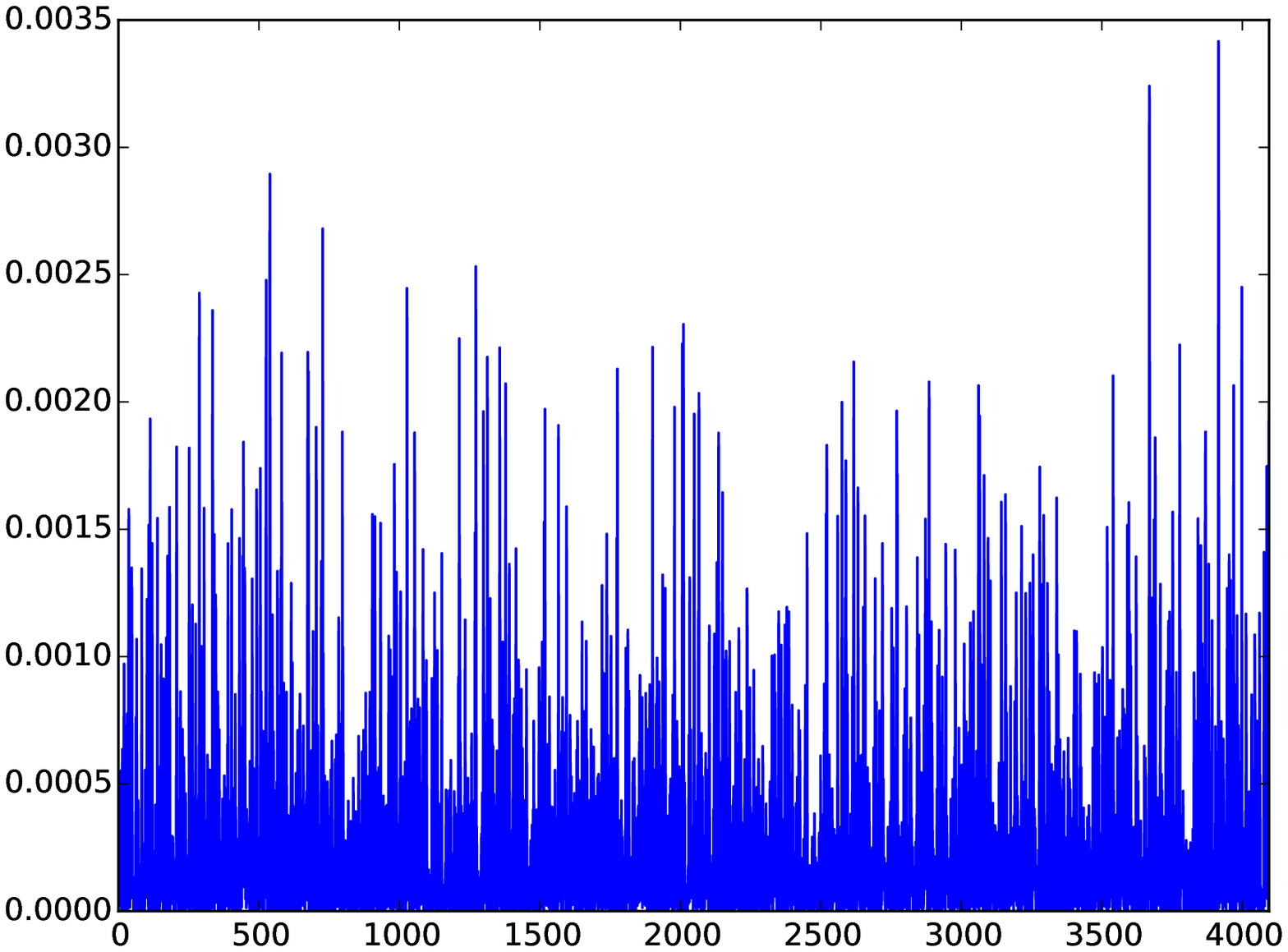}
\hspace{-0.29cm}
    \label{fig2:d3}
    \includegraphics[width=0.1\linewidth]{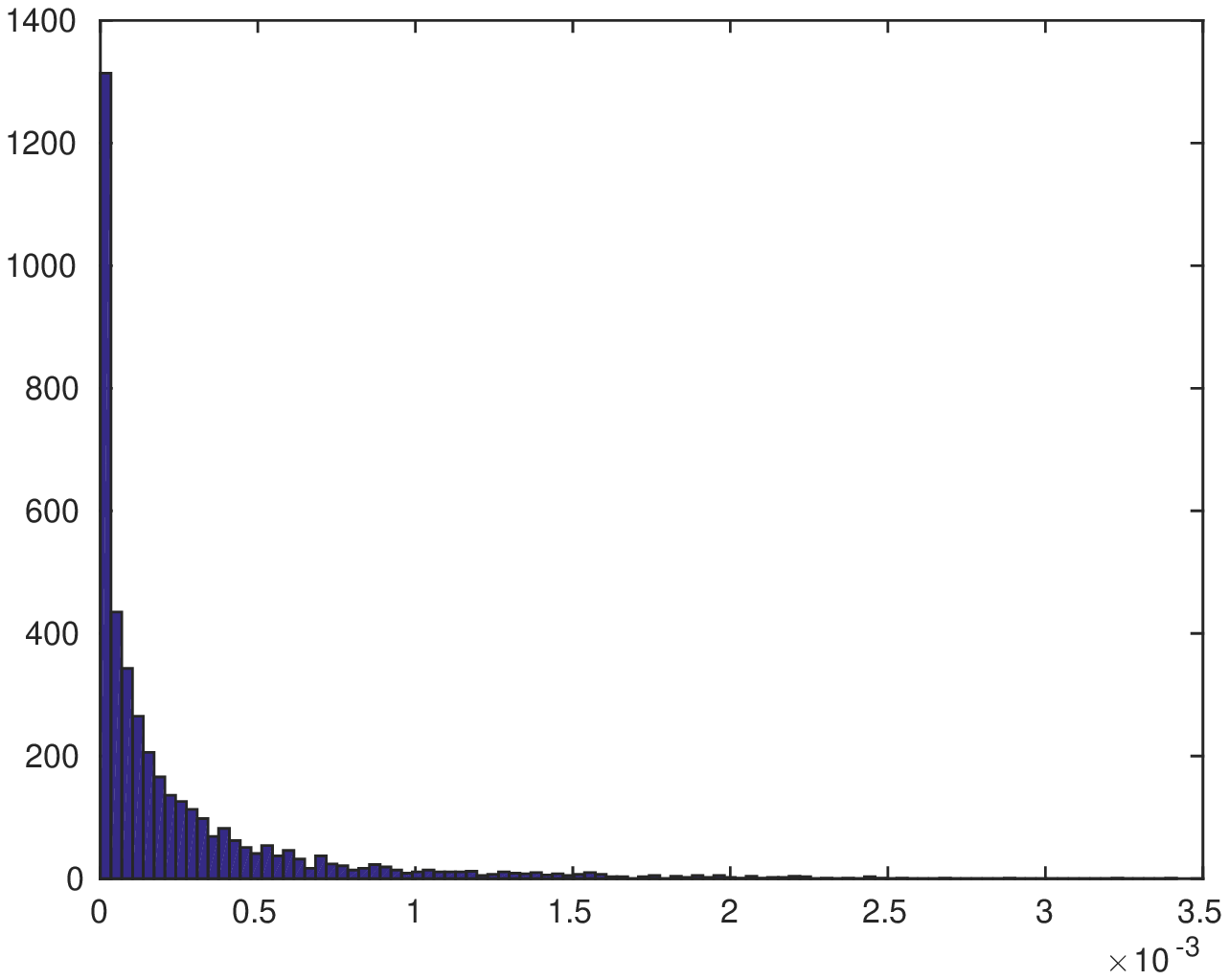}
\hspace{-0.26cm}
    \label{fig2:e3}
    \includegraphics[width=0.13\linewidth]{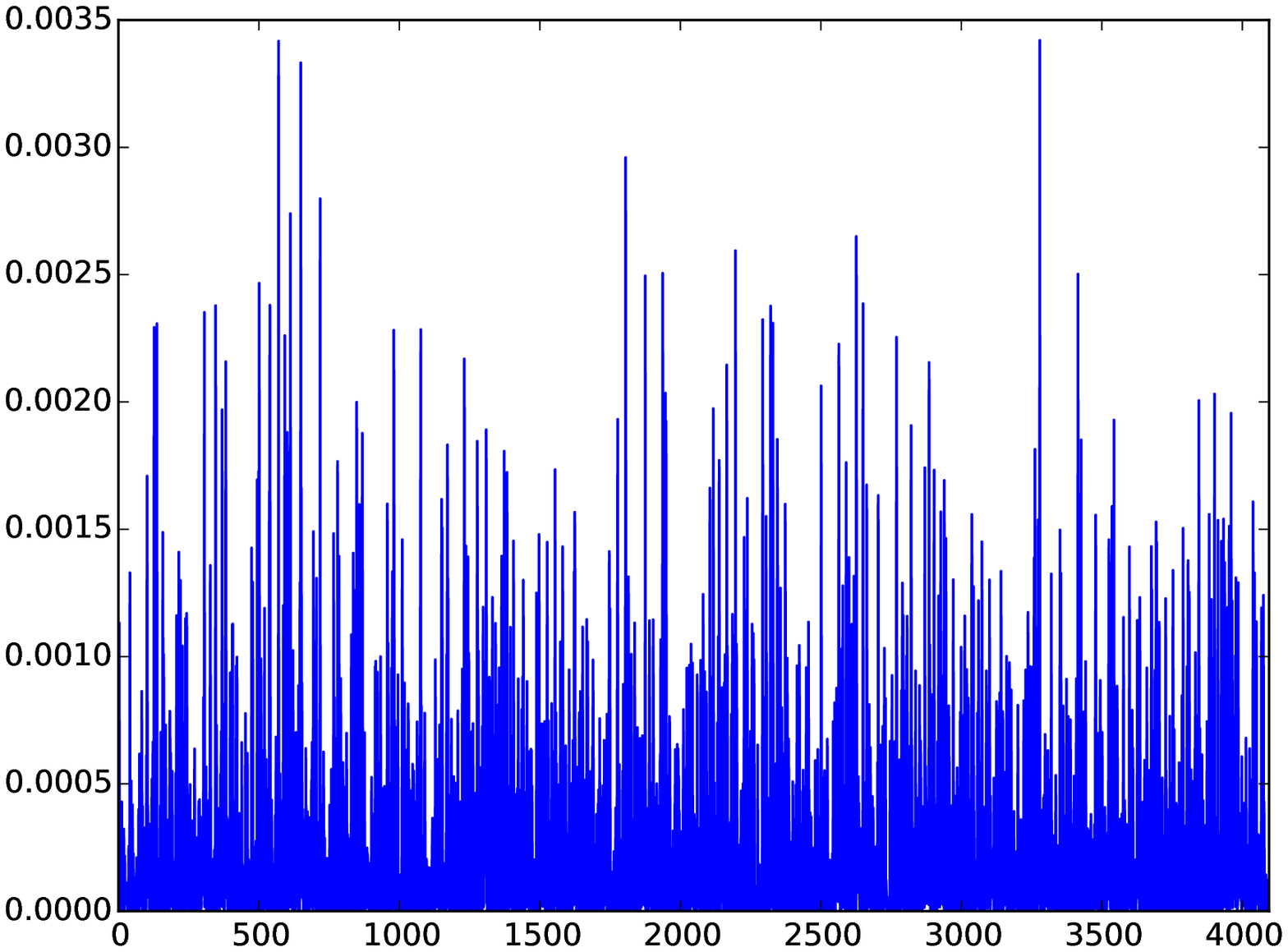}
\hspace{-0.29cm}
    \label{fig2:f3}
    \includegraphics[width=0.1\linewidth]{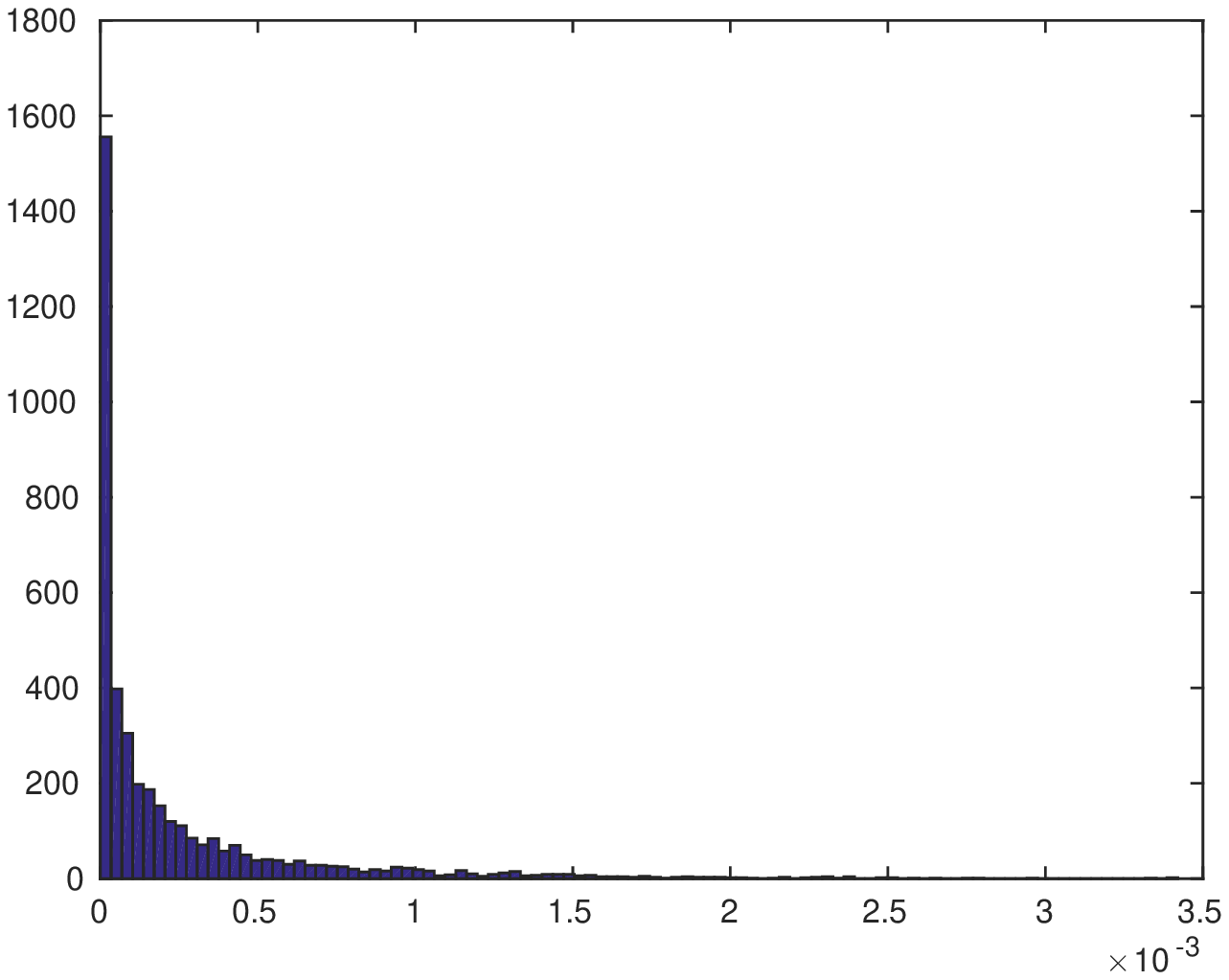}
\hspace{-0.26cm}
    \label{fig2:g3}
    \includegraphics[width=0.13\linewidth]{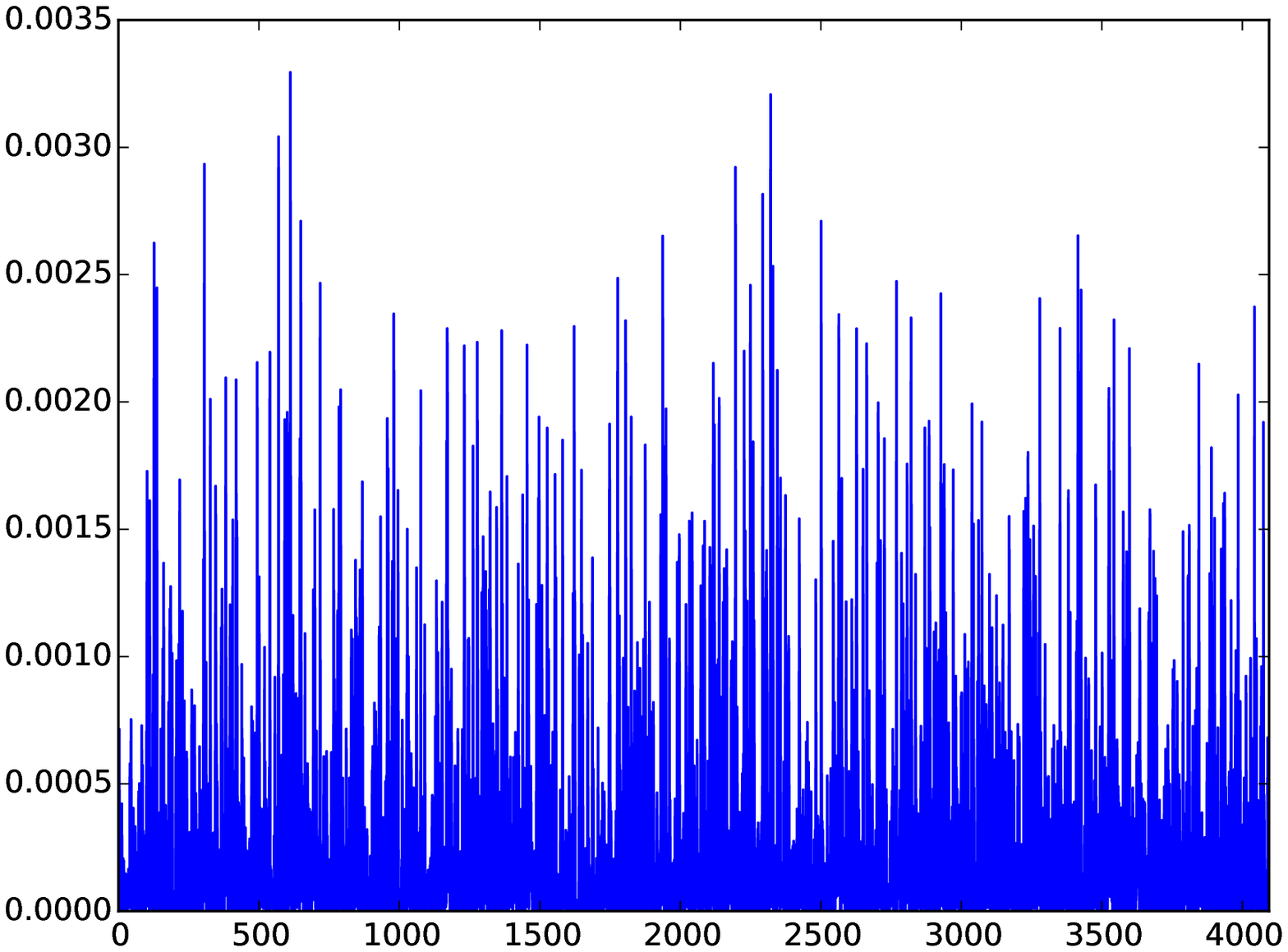}
\hspace{-0.29cm}
    \label{fig2:h3}
    \includegraphics[width=0.1\linewidth]{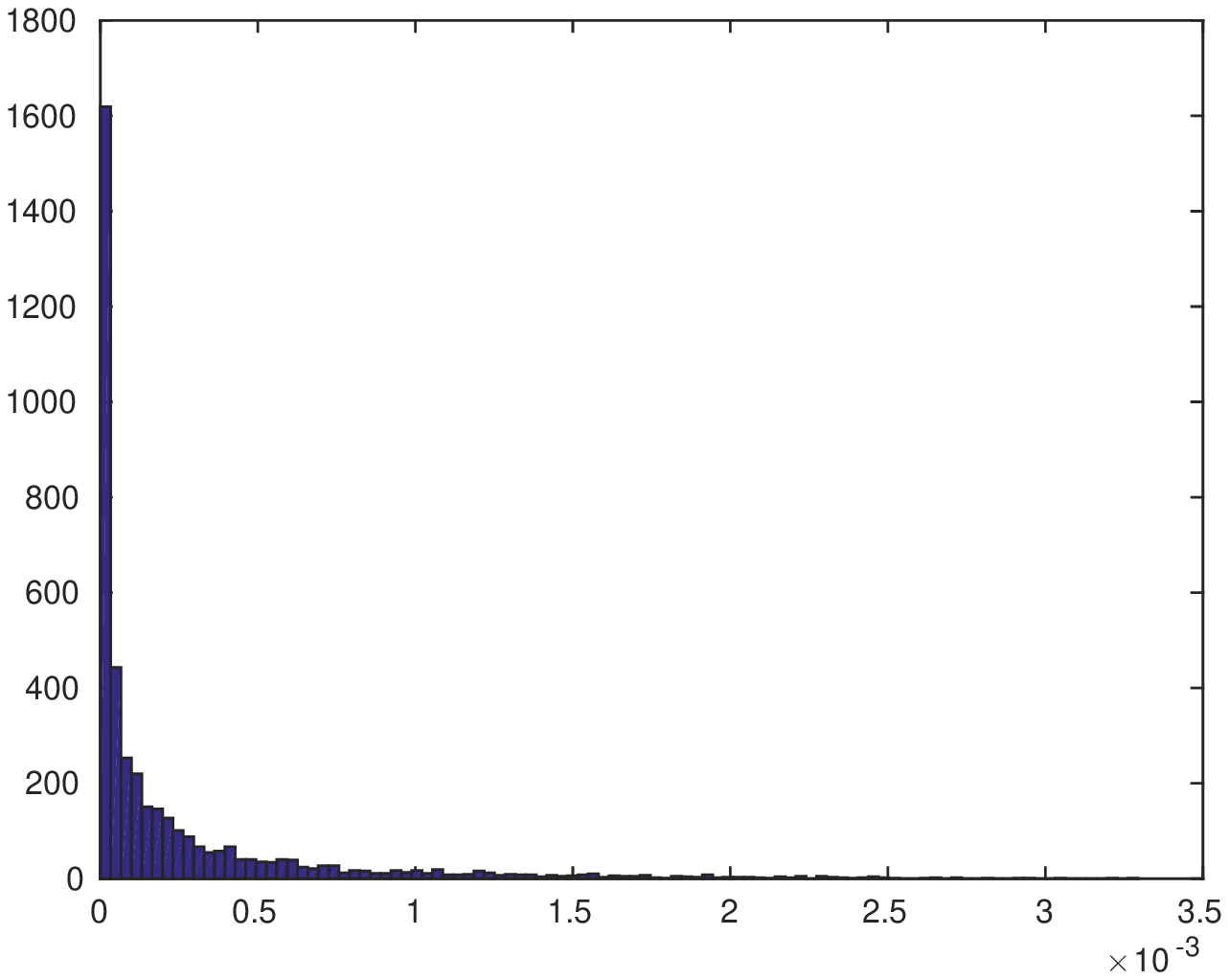}
\hspace{-0.26cm}
    \label{fig2:i3}
    \includegraphics[width=0.13\linewidth]{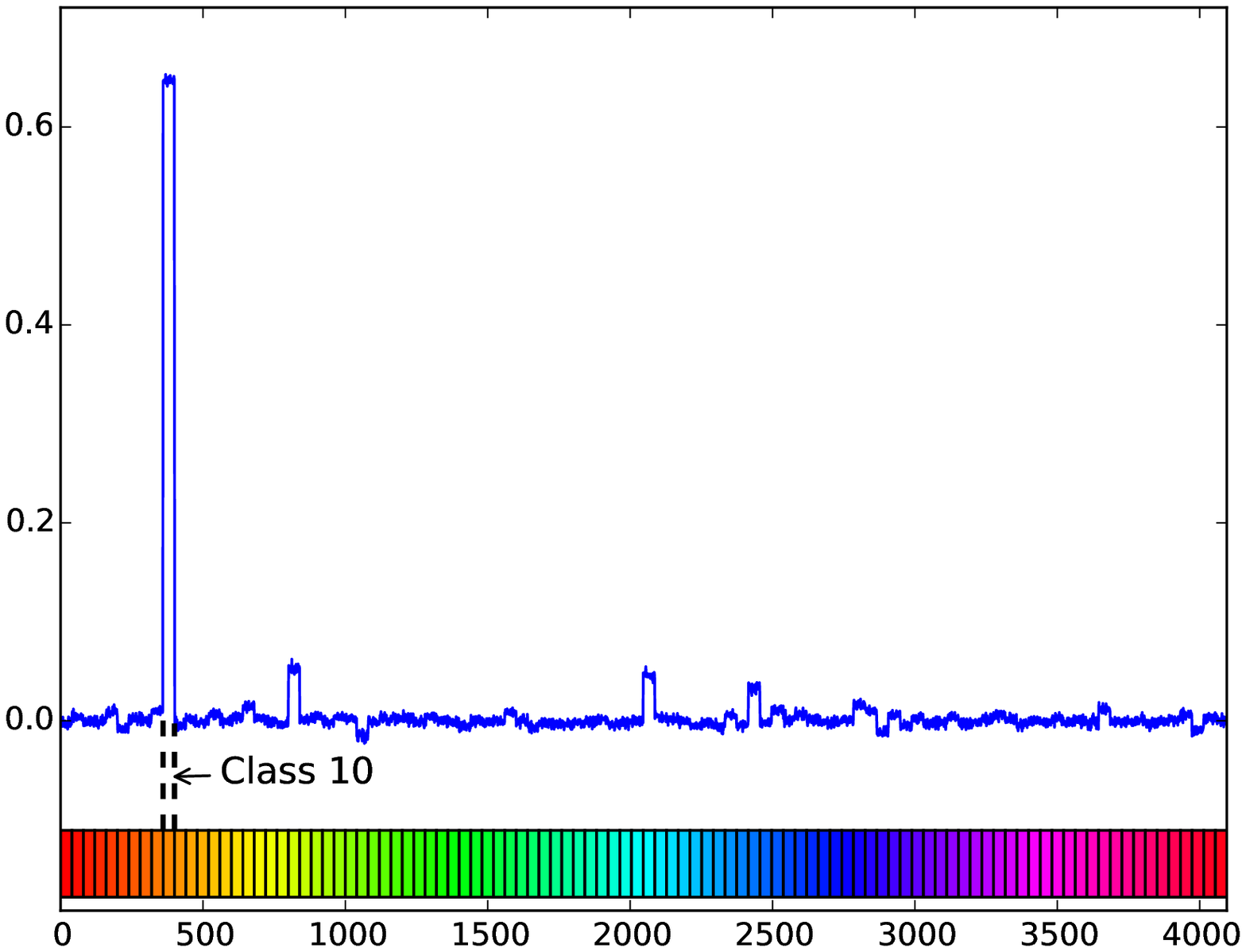}
\\
\hspace{-0.75cm}
    \label{fig2:j4}
    \includegraphics[width=0.033\linewidth]{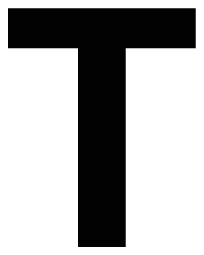}
\hspace{-0.33cm}
\subfigure[] 
{
    \label{fig2:a4}
    \includegraphics[width=0.13\linewidth]{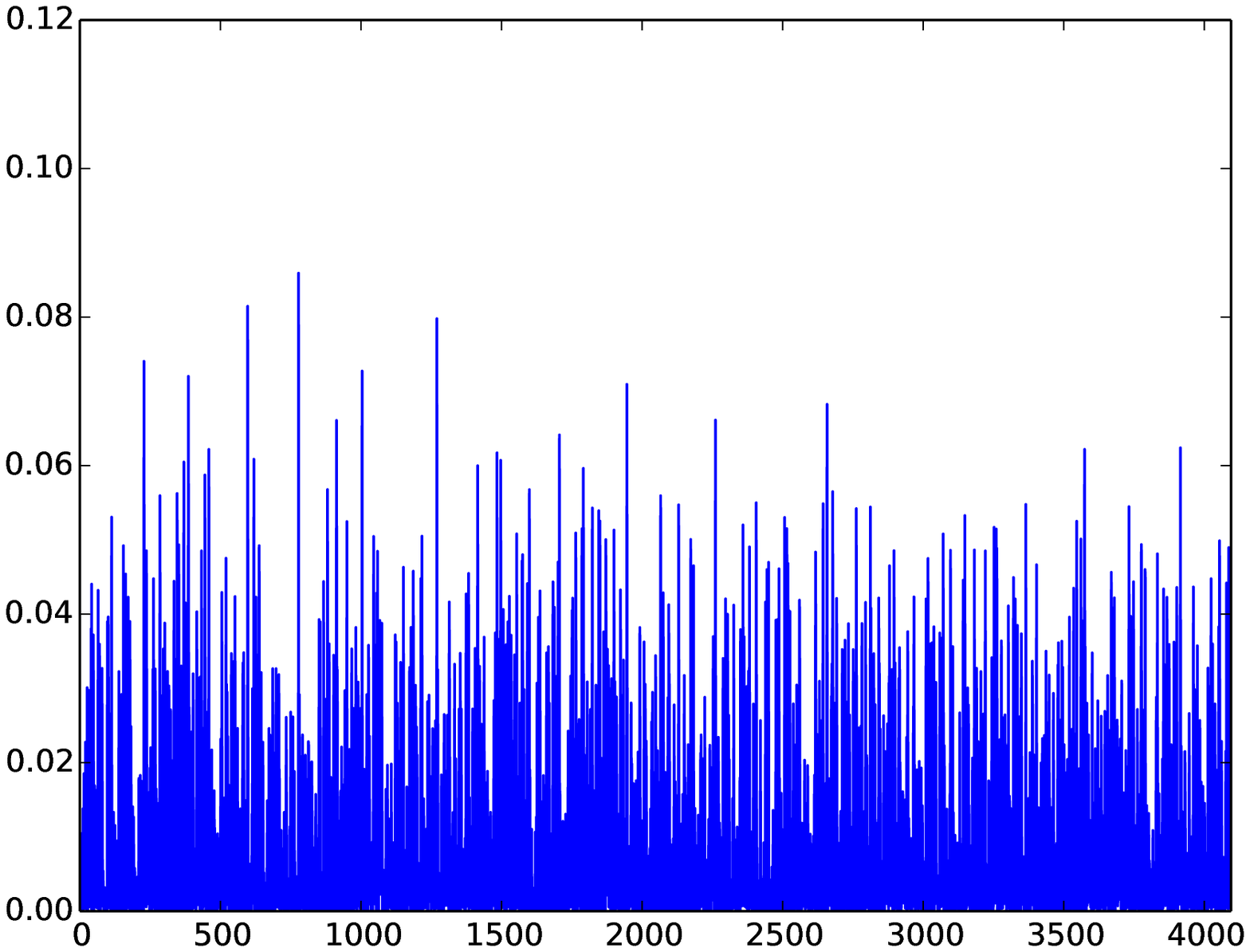}
}
\hspace{-0.44cm}
\subfigure[] 
{
    \label{fig2:b4}
    \includegraphics[width=0.1\linewidth]{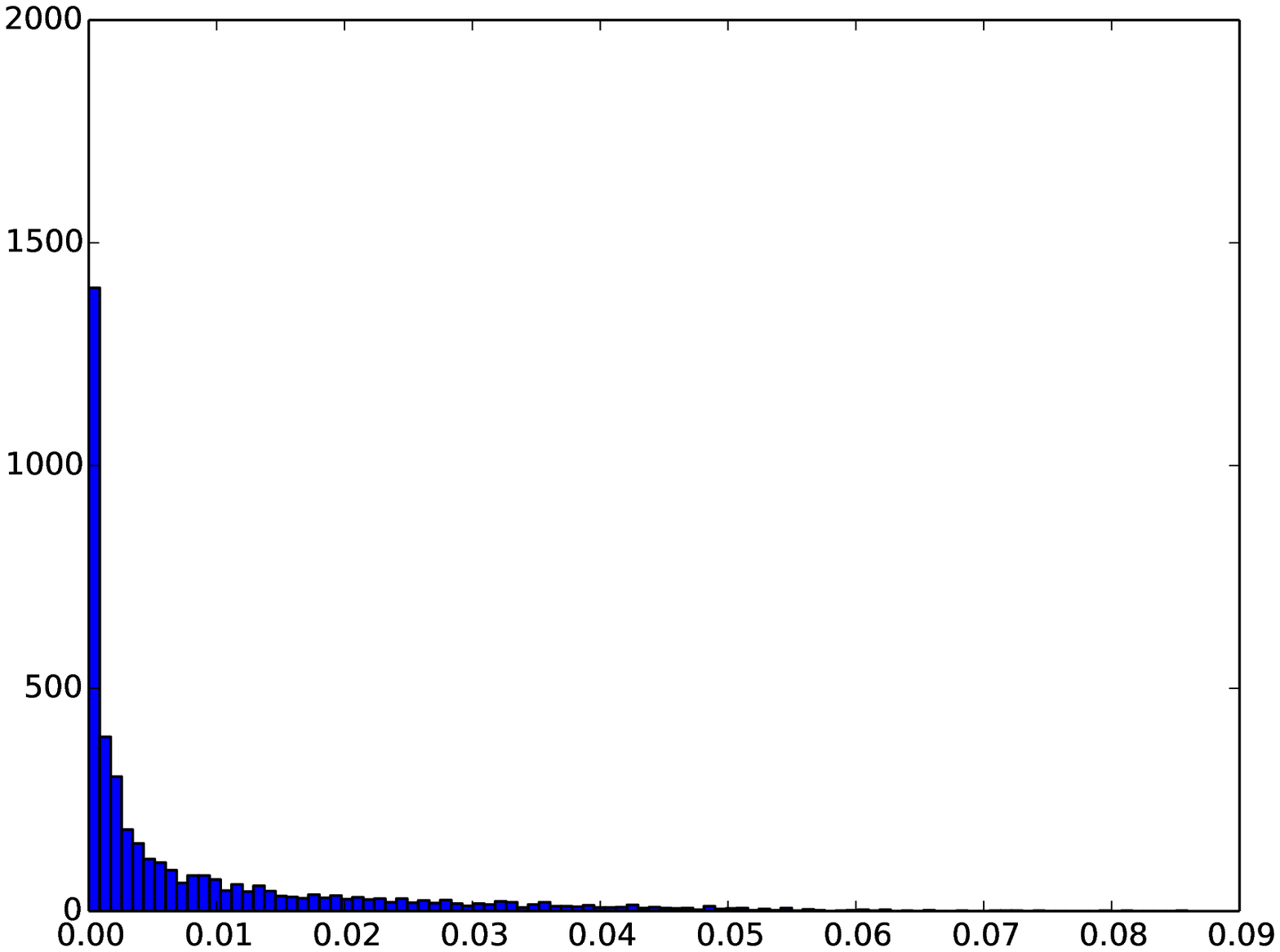}
}
\hspace{-0.44cm}
\subfigure[] 
{
    \label{fig2:c4}
    \includegraphics[width=0.13\linewidth]{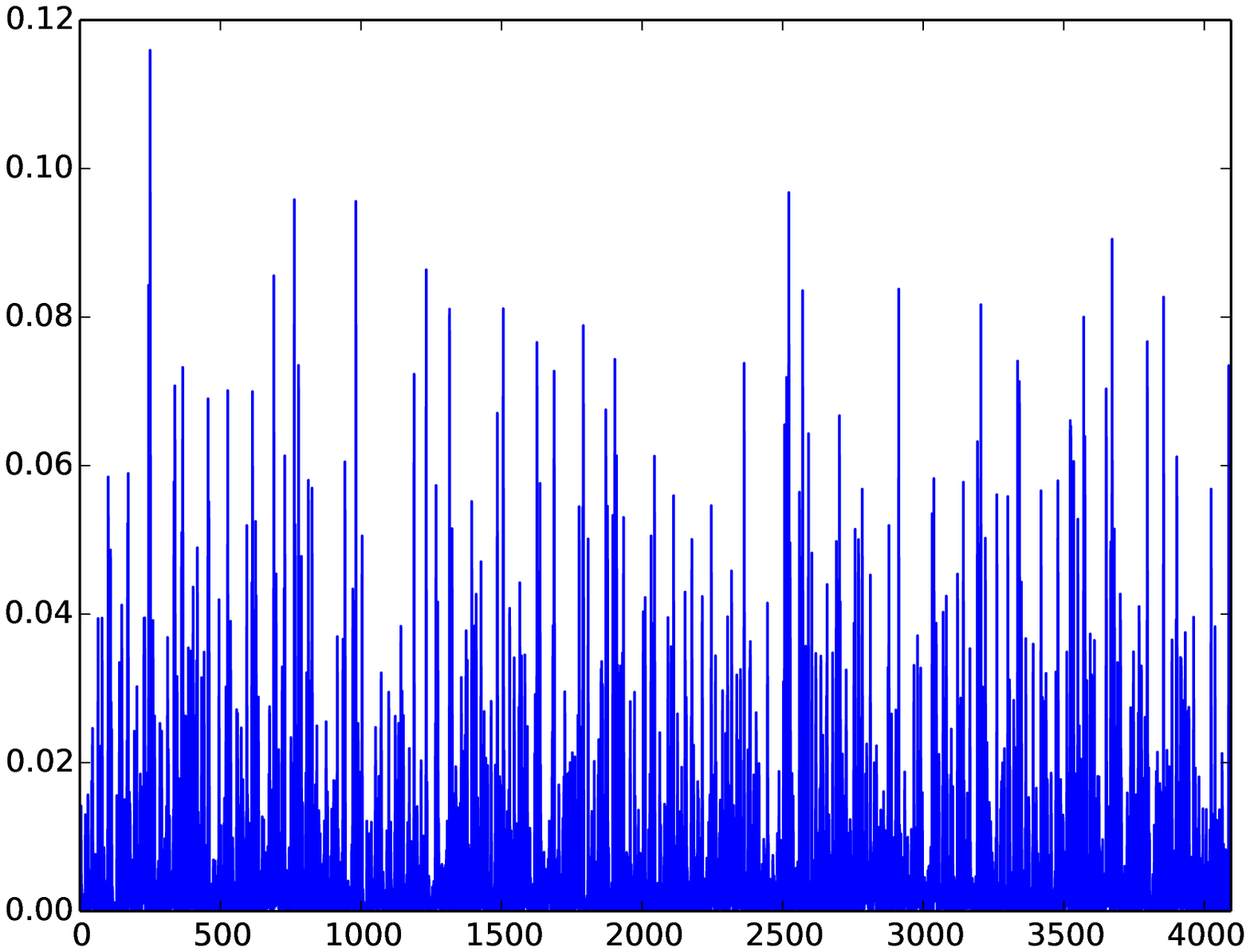}
}
\hspace{-0.44cm}
\subfigure[] 
{
    \label{fig2:d4}
    \includegraphics[width=0.1\linewidth]{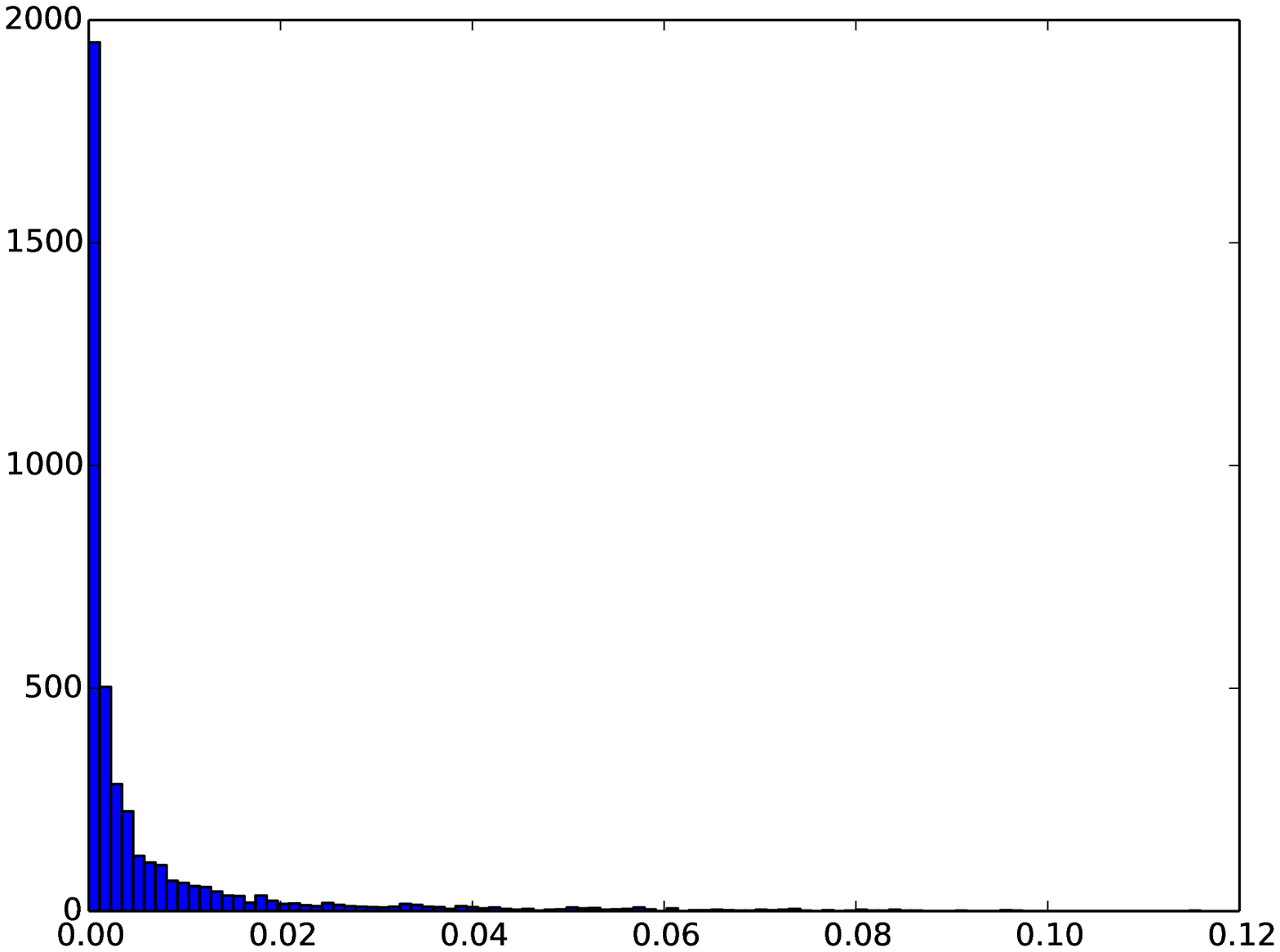}
}
\hspace{-0.44cm}
\subfigure[] 
{
    \label{fig2:e4}
    \includegraphics[width=0.13\linewidth]{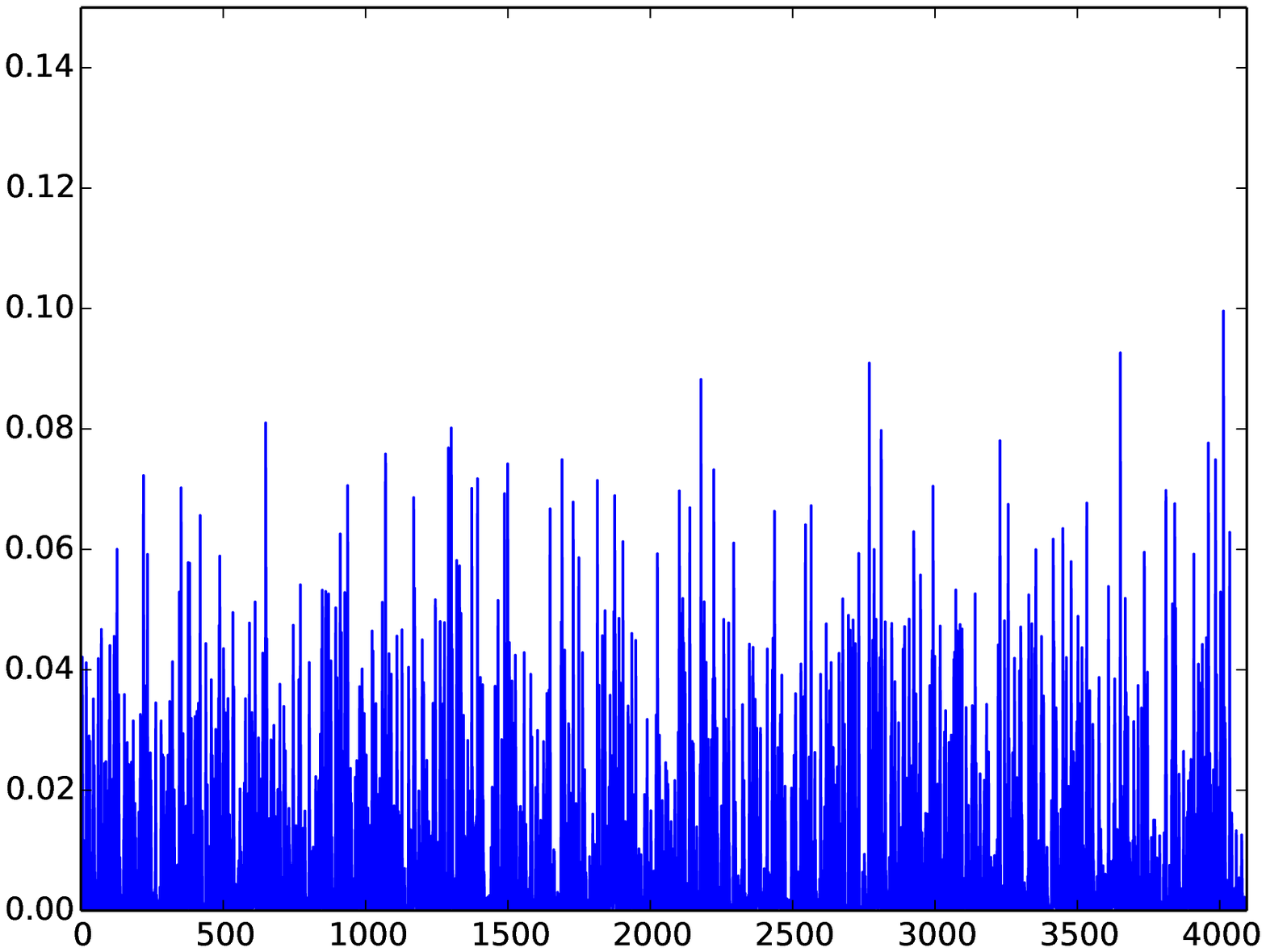}
}
\hspace{-0.44cm}
\subfigure[] 
{
    \label{fig2:f4}
    \includegraphics[width=0.1\linewidth]{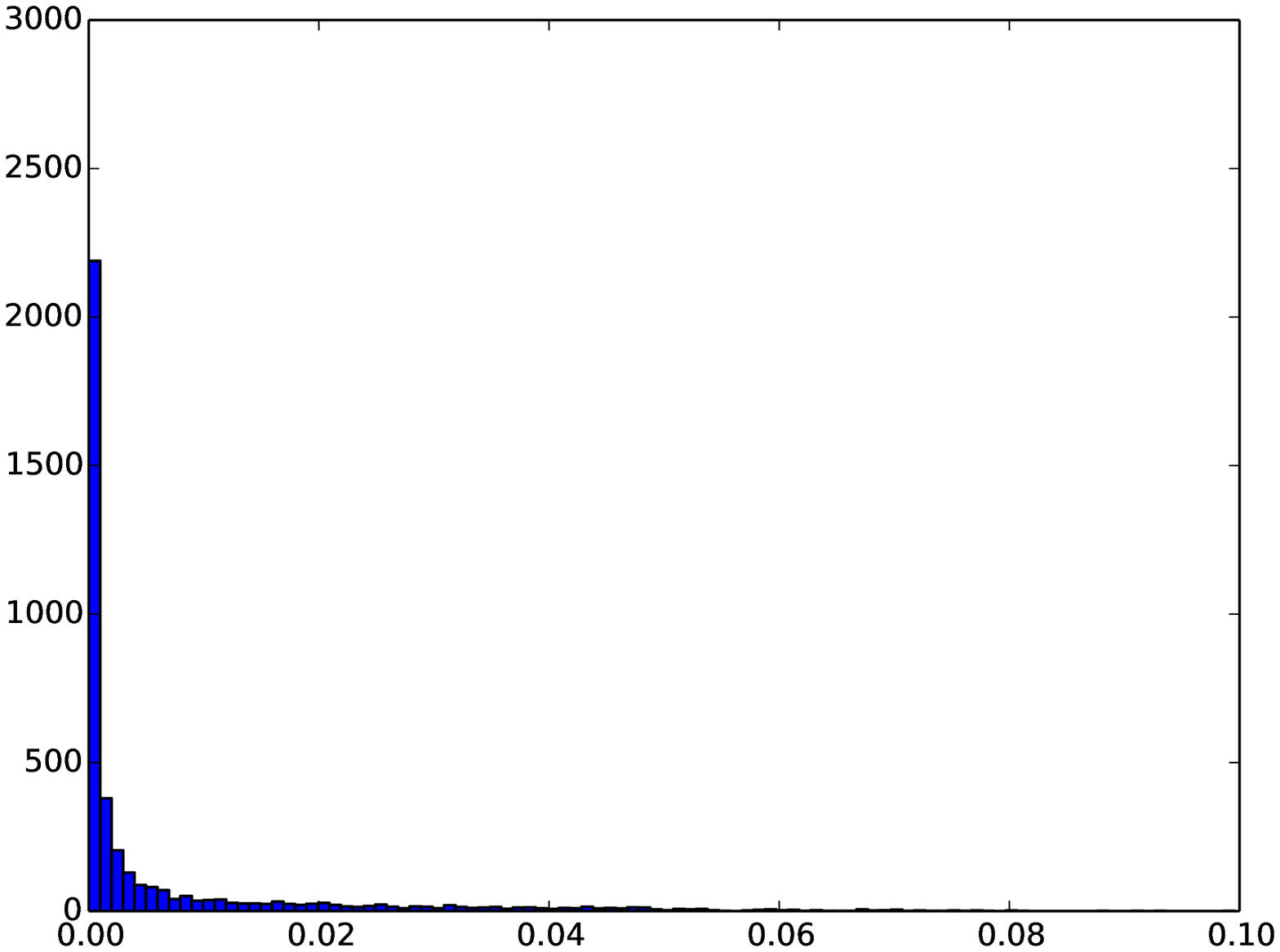}
}
\hspace{-0.44cm}
\subfigure[] 
{
    \label{fig2:g4}
    \includegraphics[width=0.13\linewidth]{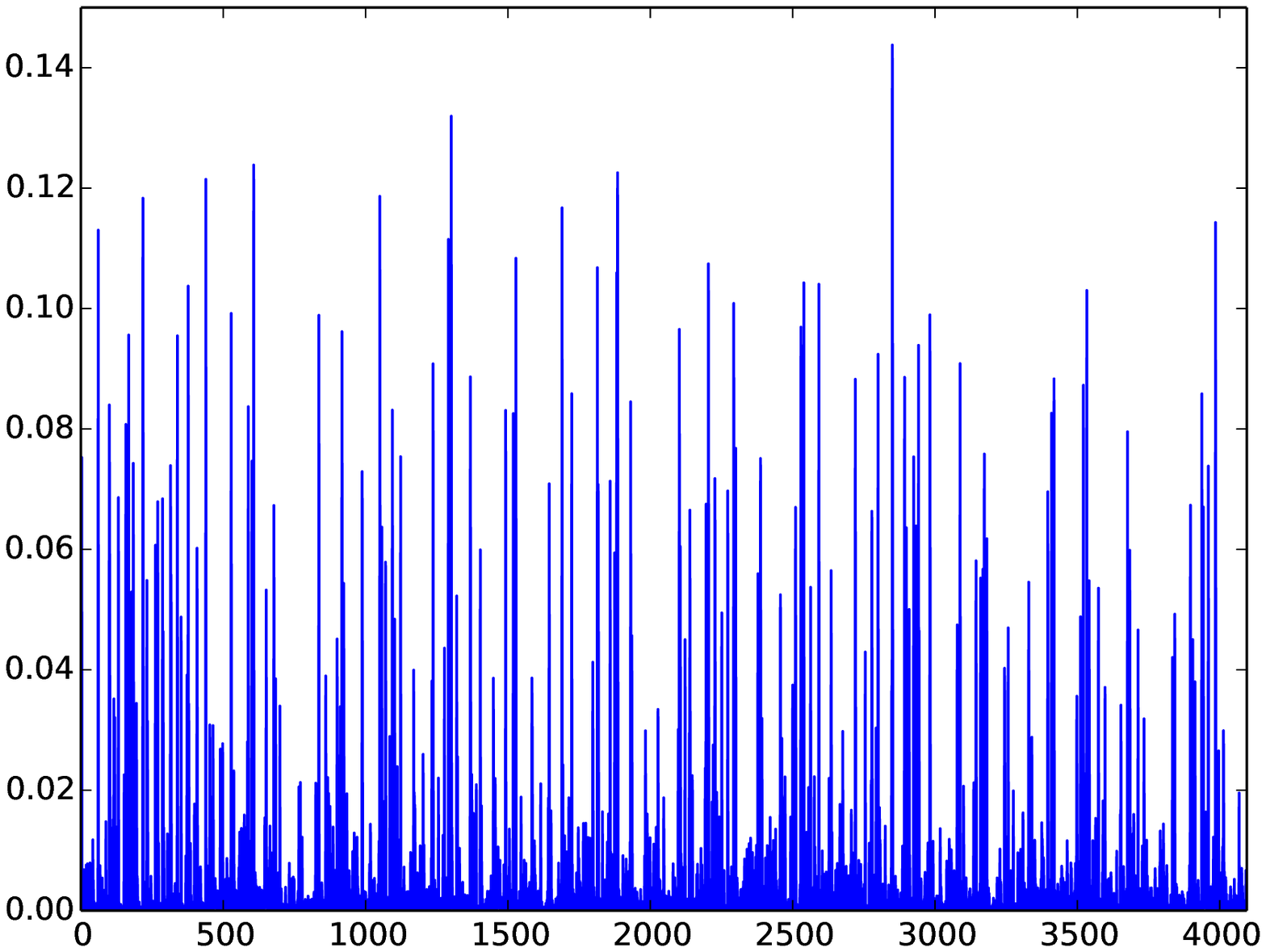}
}
\hspace{-0.44cm}
\subfigure[] 
{
    \label{fig2:h4}
    \includegraphics[width=0.1\linewidth]{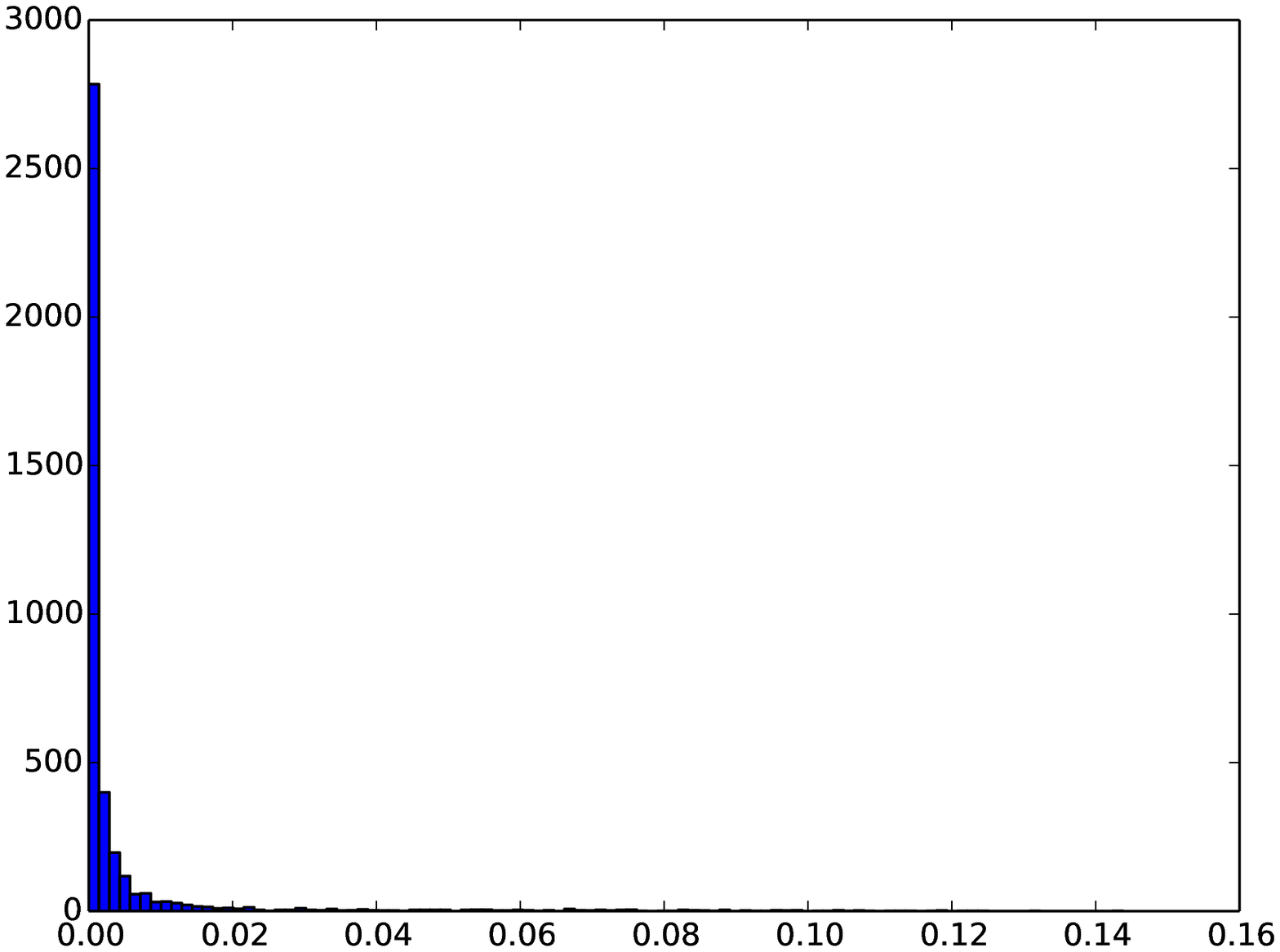}
}
\hspace{-0.45cm}
\subfigure[] 
{
    \label{fig2:i4}
    \includegraphics[width=0.13\linewidth]{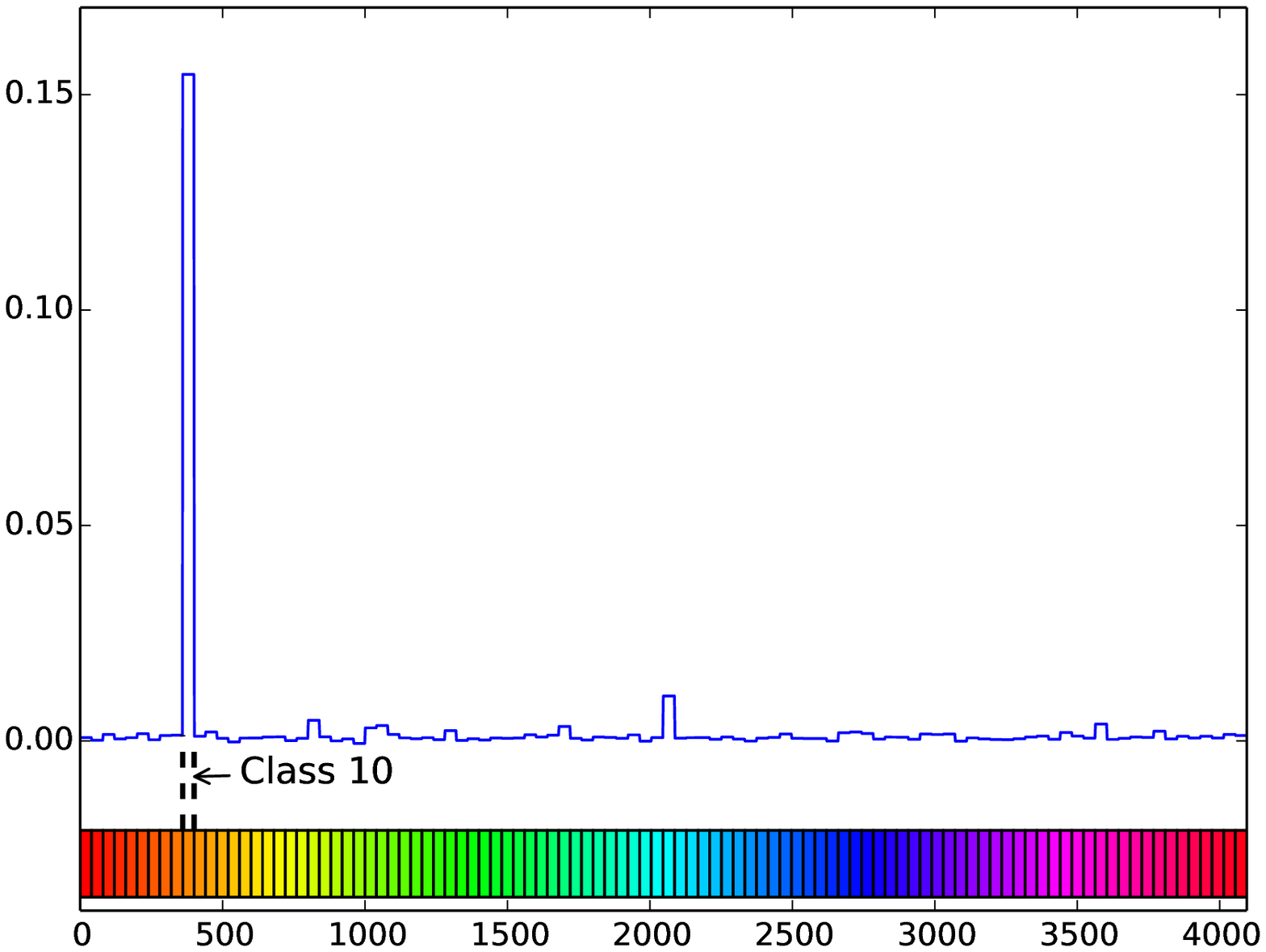}
}
\end{center}
\vspace{-0.1cm}
\caption{Examples of learned representations from layers $\text{fc}_6$, $\text{fc}_7$ and $\text{fc}_{7.5}$ using LCNN
and the baseline (VGGNet-16). Each curve indicates an average of representations for different testing videos from the same class in
the UCF101 dataset. The first two rows correspond to class 4 (Baby Crawling, 35 videos) while the third and fourth rows
correspond to class 10 (Bench Press, 48 videos). The curves in every two rows correspond to the spatial net (denoted as `S') and temporal
net (denoted as `T') in our two-stream framework for action recognition. (a) $\text{fc}_6$ representations using VGGNet-16; (b) Histograms
(with 100 bins) for representations from (a); (c) $\text{fc}_6$ representations using LCNN; (d) Histograms for
representations from (c); (e) $\text{fc}_7$ representations using VGGNet-16; (f) Histograms for representations from (e);
(g) $\text{fc}_7$ representations using LCNN; (h) Histograms for representations from (g); (i) $\text{fc}_{7.5}$
representations (i.e. transformed $\text{fc}_7$ representations) using LCNN. The entropy values for representations from (a)(c)(e)(g) are computed as: (11.32, 11.42, 11.02, 10.75),
(11.2, 11.14, 10.81, 10.34), (11.08, 11.35, 10.67, 10.17), (11.02, 10.72, 10.55, 9.37). LCNN can generate lower-entropy representations for each class compared to VGGNet-16.
Each color from the color bars in (i) represents one class for a subset of neurons. The black dashed lines indicate that the curves are highly peaked in one class. The figure is best viewed in color and 600\% zoom in.}
\label{fig2}
\vspace{-0.5cm}
\end{figure*}

\subsection{Formulation} \label{sec4_2}

The overall objective function of LCNN is a combination of the discriminative representation error
at late hidden layers and the classification error at the output layer:
\begin{align}
L &= L_c + \alpha L_r \label{eq10}
\end{align}
where $L_c$ in Equation (\ref{eq2}) is the classification error at the output layer, $L_r$
is the discriminative representation error in Equation (\ref{eq6}) and will be discussed in detail below, and
$\alpha$ is a hyper parameter balancing the two terms.

Suppose we want to add supervision to the $l^{\text{th}}$ layer. Let $(\mathbf{x}, y)$ denote a training sample and
$\mathbf{x}^{(l)} \in \mathbb{R}^{N_l}$ be the corresponding representation produced by the $l^{\text{th}}$ layer,
which is defined by the activations of $N_l$ neurons in that layer. Then the discriminative representation error is defined to be
the difference between the transformed representation $\mathbf{A}^{(l)}\mathbf{x}^{(l)}$ and the ideal discriminative representation $\mathbf{q}^{(l)}$:
\begin{equation} \label{eq6}
L_r = L_r(\mathbf{x}^{(l)}, y, \mathbf{A}^{(l)}) = \lVert \mathbf{q}^{(l)} - \mathbf{A}^{(l)}\mathbf{x}^{(l)} \rVert_2^2,
\end{equation}
where $\mathbf{A}^{(l)} \in \mathbb{R}^{N_l\times N_l}$ is a linear
transformation matrix, and the binary vector $\mathbf{q}^{(l)} = [q^{(l)}_{1}, \ldots,q^{(l)}_{j},\ldots, q^{(l)}_{N_l}]^{\rm{T}} \in \{0, 1\}^{N_l}$
denotes the ideal discriminative
representation which indicates the ideal activations of neurons ($j$ denotes the index of neuron, \textit{i.e.} the index of
feature dimension). Each neuron is associated with a certain class
label and, ideally, only activates to samples from that class. Therefore, when a sample is from Class $c$,
$q^{(l)}_{j} = 1$ if and only if the $j^{\rm{th}}$ neuron is assigned to Class $c$, and neurons associated to other
classes should not be activated so that the corresponding entry in $\mathbf{q}^{(l)}$ is zero. Notice that
$\mathbf{A}^{(l)}$ is the only parameter needed to be learned,
while $\mathbf{q}^{(l)}$ is pre-defined based on label information from training data.

Suppose we have a batch of six training samples $\{\mathbf{x}_1,\mathbf{x}_2, \ldots,\mathbf{x}_6\}$ and the class labels $\mathbf{y} = [y_1, y_2, \ldots, y_6] = [1, 1,
2, 2, 3, 3]$. Further assume that the $l^{\text{th}}$ layer has 7 neurons $\{d_1,d_2,\ldots,d_7\}$ with $\{d_1,d_2$\} associated
with Class 1, $\{d_3, d_4, d_5\}$ Class 2, and $\{d_6, d_7\}$ Class 3. Then the ideal discriminative representations for these
six samples are given by:
\begin{align}
{\mathbf{Q}}^{(l)}
= \left[
\begin{array} {cccccc}
1 & 1 & 0 & 0 & 0 & 0 \\
1 & 1 & 0 & 0 & 0 & 0 \\
0 & 0 & 1 & 1 & 0 & 0 \\
0 & 0 & 1 & 1 & 0 & 0 \\
0 & 0 & 1 & 1 & 0 & 0 \\
0 & 0 & 0 & 0 & 1 & 1 \\
0 & 0 & 0 & 0 & 1 & 1
\end{array}
\right], \label{eq8}
\end{align}
where each column is an ideal discriminative representation corresponding to a training sample. The ideal
representations ensured that the input signals from the same class have similar representations while those from different
classes have dissimilar representations.

The discriminative representation error (\ref{eq6}) forces the learned representation to approximate the ideal
discriminative representation, so that the resulting neurons have the label consistency property \cite{lcksvd}, \textit{i.e.} the class distributions of each neuron~\footnote{Similar to computing the class distributions for dictionary items in~\cite{Qiu11}, the class distributions of each neurons from the $l^{\text{th}}$ layer can be derived by measuring their activations $\mathbf{x}^{(l)}$ over input signals corresponding to different classes.} from layer $l$ are extremely peaked in one class. In addition, with more discriminative representations, the classifier, especially linear classifiers, at the output layer can achieve better performance. This is because the discriminative property of $\mathbf{x}^{(l)}$ is very important for the performance of a linear classifier.

An example of the LCNN architecture is shown in Figure \ref{fig1}. The linear
transformation is implemented as a fully-connected layer. We refer it as `Transformed Representation Layer'. We create a new `Ideal Representation Layer' which transforms a class label into the corresponding binary vector $\mathbf{q}^{(l)}$; then we feed the outputs of these two layers into the Euclidean loss layer.

In our experiments, we allocate the neurons in the late hidden layer to each class as follows: assuming $N_l$ neurons in that layer and $m$ classes,  we first allocate $\lfloor N_l/m \rfloor$ neurons to each class and then allocate the remaining ($N_l - m$$\lfloor N_l/m \rfloor$) neurons to the top ($N_l - m$$\lfloor N_l/m \rfloor$) classes with high intra-class appearance variation. Therefore each neuron in the late hidden layer is associated with a category label, but an input signal of a category certainly can (and does) use all neurons (learned features), as the representations in Figure~\ref{fig2:i4} illustrate, \textit{i.e.} sharing features between categories is not prohibited.

\subsection{Network Training} \label{sec4_3}

LCNN is trained via stochastic gradient descent. We need to compute the gradients of $L$ in Equation
(\ref{eq10}) w.r.t. all the network
parameters $\{\mathbf{W}, \mathbf{A}^{(l)}\}$. Compared with standard CNN, the difference lies in two gradient terms,
\textit{i.e.} $\frac {\partial L} {\partial \mathbf{x}^{(l)}}$ and $\frac {\partial L} {\partial \mathbf{A}^{(l)}}$, since
$\mathbf{x}^{(l)}$ and $\mathbf{A}^{(l)}$ are the only parameters which are related to the newly added
discriminative error $L_r(\mathbf{x}^{(l)}, y, \mathbf{A}^{(l)})$ and the other parameters act independently from it.

It follows from Equations (\ref{eq10}) and (\ref{eq6}) that
\begin{align}
\frac {\partial L} {\partial \mathbf{x}^{(i)}} &=
\begin{cases}
\frac {\partial L_c} {\partial \mathbf{x}^{(i)}},\quad i \neq l \\
\frac {\partial L_c} {\partial \mathbf{x}^{(l)}} + 2\alpha(\mathbf{A}^{(l)} \mathbf{x}^{(l)} - \mathbf{q}^{(l)})^{\rm
T} \mathbf{A}^{(l)}, ~i = l
\end{cases}\\
\frac {\partial L} {\partial \mathbf{W}^{(i)}} &= \frac {\partial L_c} {\partial \mathbf{W}^{(i)}}, \quad \forall i \in \{1, 2, ..., n\}\\
\frac {\partial L} {\partial \mathbf{A}^{(l)}} &= 2\alpha(\mathbf{A}^{(l)}\mathbf{x}^{(l)} -
\mathbf{q}^{(l)})\mathbf{x}^{(l) {\rm T}},
\end{align}
where $\frac {\partial L_c} {\partial \mathbf{x}^{(i)}}$ and $\frac {\partial L_c} {\partial \mathbf{W}^{(i)}}$ are computed by
Equations (\ref{eq4}) and (\ref{eq5}), respectively.

\begin{figure}[t]
\begin{center}
\includegraphics[width=0.9\linewidth]{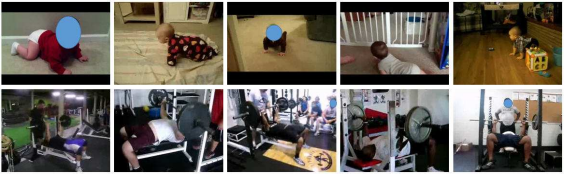}
\end{center}
\vspace{-0.1cm}
\caption{Class 4 (BabyCrawling) and class 10 (BenchPress) samples from the UCF101 action dataset.}
\label{fig4}
\end{figure}

\section{Experiments}\label{sec5}

We evaluate our approach on two action recognition datasets: UCF101 \cite{ucf101} and THUMOS15 \cite{thumos15}, and three object category datasets: Cifar-10
\cite{cifar10}, ImageNet \cite{imagenet} and Caltech101 \cite{caltech101}. Our implementation of LCNN is based on the CAFFE toolbox~\cite{caffe}.

To verify the effectiveness of our label consistency module, we train LCNN in two ways: (1) We use the discriminative representation error loss $L_r$ only; (2) We
use the combination of $L_r$ and the softmax classification error loss $L_c$ as in Equation (\ref{eq10}). We refer to the networks trained in these ways as `LCNN-1' and `LCNN-2', respectively.
The baseline is to use the softmax classification error loss $L_c$ only during network training.
We refer to it as `baseline' in the following. Note that the baseline and LCNN are trained with the same parameter setting and initial model in all our experiments.

For action and object recognition, we introduce two classification approaches here: (1) \textbf{argmax}: we follow the standard CNN
practice of taking the class label corresponding to the
maximum prediction score;  (2) \textbf{$k$-NN}: We use the transformed representation $\mathbf{A}^{(l)}\mathbf{x}^{(l)}$ to represent an image, video frame or optical flow field and then use a simple $k$-NN classifier. LCNN-1 always uses `$k$-NN' for classification while LCNN-2 can use either `argmax' or `$k$-NN' to do classification.


\begin{table}[t]
\centering
\begin{tabular}{|c|c|c|c|}
\hline
Network Architecture & Spatial & Temporal & Both\\
\hline
ClarifaiNet~\cite{twostream1} & 72.7 & 81 & 87 \\
VGGNet-19~\cite{twostream3} & 75.7 & 78.3 & 86.7 \\
VGGNet-16~\cite{twostream2} & 79.8 & 85.7 & 90.9 \\
VGGNet-16*~\cite{twostream2} & - & 85.2 & - \\
baseline &  77.48 & 83.71 & - \\
LCNN-1 & 80.1 & 85.59 & 89.87\\
LCNN-2 (argmax) & 80.7 & 85.57 & 91.12\\
LCNN-2 ($k$-NN) & 81.3 & 85.77 & 89.84\\
\hline
\end{tabular}
\vspace{0.2cm}
\caption{\label{tb2}Classification performance with different two-stream CNN approaches on the UCF101 dataset (split-1). The results
of~\cite{twostream1,twostream2,twostream3} are copied from their original papers. The VGGNet-16* result is
obtained by testing the model shared by~\cite{twostream2}. The `baseline' are the results of running
the two-stream CNN implementation provided by~\cite{twostream2}, where the VGGNet-16 architecture is used for each stream.
LCNN and baseline are trained with the same parameter setting and initial model. The only difference between LCNN-2 and the baseline is
that we add explicit supervision to $\text{fc}_7$ layer for LCNN-2. For LCNN-1, we remove the softmax layer from the baseline network
but add explicit supervision to $\text{fc}_7$ layer.}
\end{table}

\begin{table}[t]
\centering
\begin{tabular} {|c|c|c|c|}
\hline
Method & Acc. (\%)  & Method & Acc. (\%)\\
\hline
Karpathy~\cite{karpathy14} & 65.4 & Wang~\cite{idt} & 85.9\\
Donahue~\cite{Donahue15} & 82.9 & Lan~\cite{Lan15} & 89.1 \\
Ng~\cite{joe15} & 88.6 & Zha~\cite{shengxin15} & 89.6\\
LCNN-2 (argmax) & 91.12 & & \\
\hline
\end{tabular}
\vspace{0.2cm}
\caption{\label{tb1}Recognition performance comparisons with other state-of-the-art approaches on the UCF101 dataset. The results of ~\cite{karpathy14,idt,Donahue15,Lan15,joe15,shengxin15} are copied from their original papers.}
\end{table}

\begin{figure}[t]
\begin{center}
\centering
\subfigure[]
{
\label{fig3:a}
\includegraphics[width=0.7\linewidth]{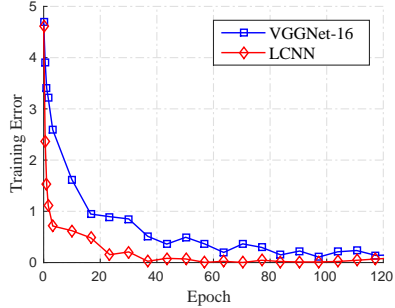}
}\\
\subfigure[]
{
\label{fig3:b}
\includegraphics[width=0.7\linewidth]{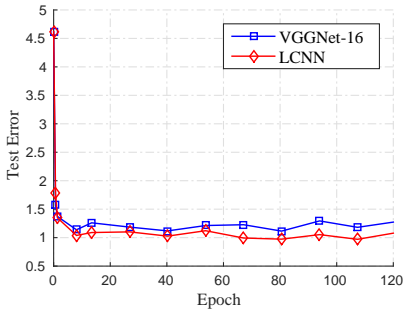}
}
\end{center}
\vspace{-0.1cm}
\caption{Training and testing errors of spatial net trained by LCNN-2 and the baseline (VGGNet-16) on the UCF101 dataset. (a) Training error comparison; (b) Testing error comparison.}
\label{fig3}
\end{figure}

\begin{figure}
\begin{center}
\includegraphics[width=0.7\linewidth]{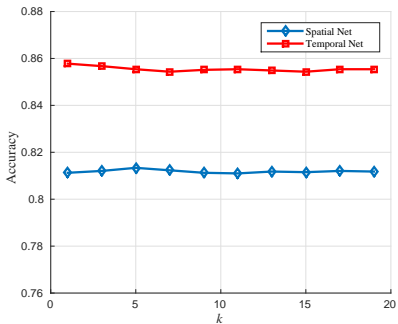}
\end{center}
\vspace{-0.1cm}
\caption{Effects of parameter selection of $k$-NN neighborhood size $k$ on the classification accuracy performances on the UCF101 dataset. The spatial and temporal nets trained by LCNN-2 are not sensitive to the selection of $k$.}
\label{fig5}
\end{figure}

\subsection{Action Recognition} \label{sec5_2}

\subsubsection{UCF101 Dataset} \label{sec5_2_1}

The UCF101 dataset~\cite{ucf101} consists of $13,320$ video clips from 101 action classes, and every class has more than $100$ clips. Some video examples from class 4 and class 10 are given in Figure~\ref{fig4}. In terms of evaluation, we use the standard split-1 train/test setting to evaluate our approach. Split-1 contains around $10,000$ clips for training and the rest for testing.

We choose the popular two-stream CNN as in~\cite{twostream1,twostream2,twostream3} as our basic network
architecture for action recognition. It consists of a spatial net taking video frames as input and a temporal net taking 10-frame stacking of optical flow fields. Late fusion is conducted on the outputs of the two streams and generates the final prediction score. During testing, we sample 25 frames (images or optical flow fields) from a video as in~\cite{twostream1} for spatial and temporal nets. The class scores for a testing video is obtained by averaging the scores across sampled frames. In our experiments, we fuse spatial and temporal net prediction scores using a simple weighted average rule, where the weight is set to 2 for temporal net and 1 for spatial net.

\begin{figure}[t]
\begin{center}
\centering
\subfigure[]
{
\label{fig6:a}
\includegraphics[width=0.43\linewidth]{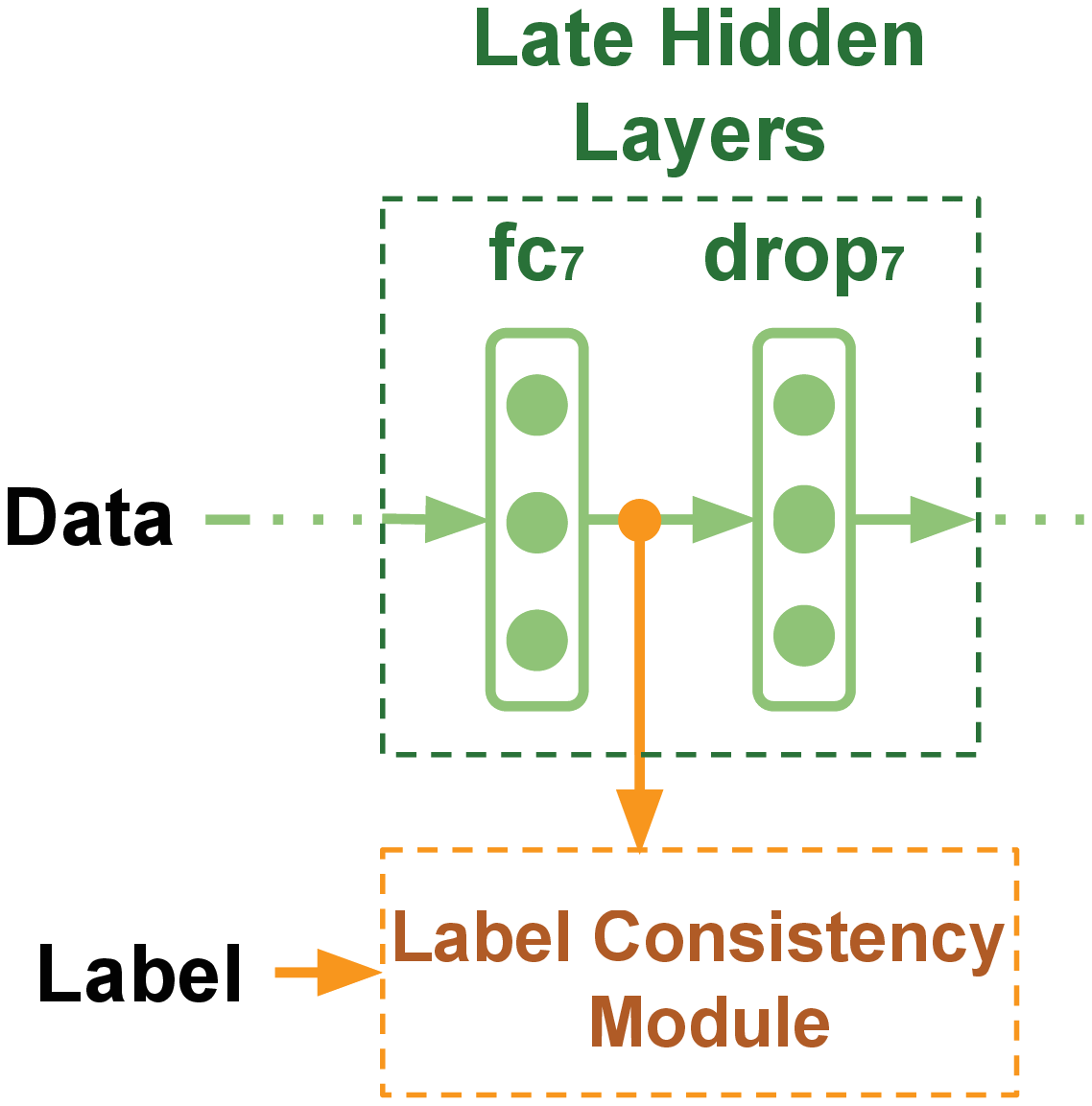}
}
\subfigure[]
{
\label{fig6:b}
\includegraphics[width=0.43\linewidth]{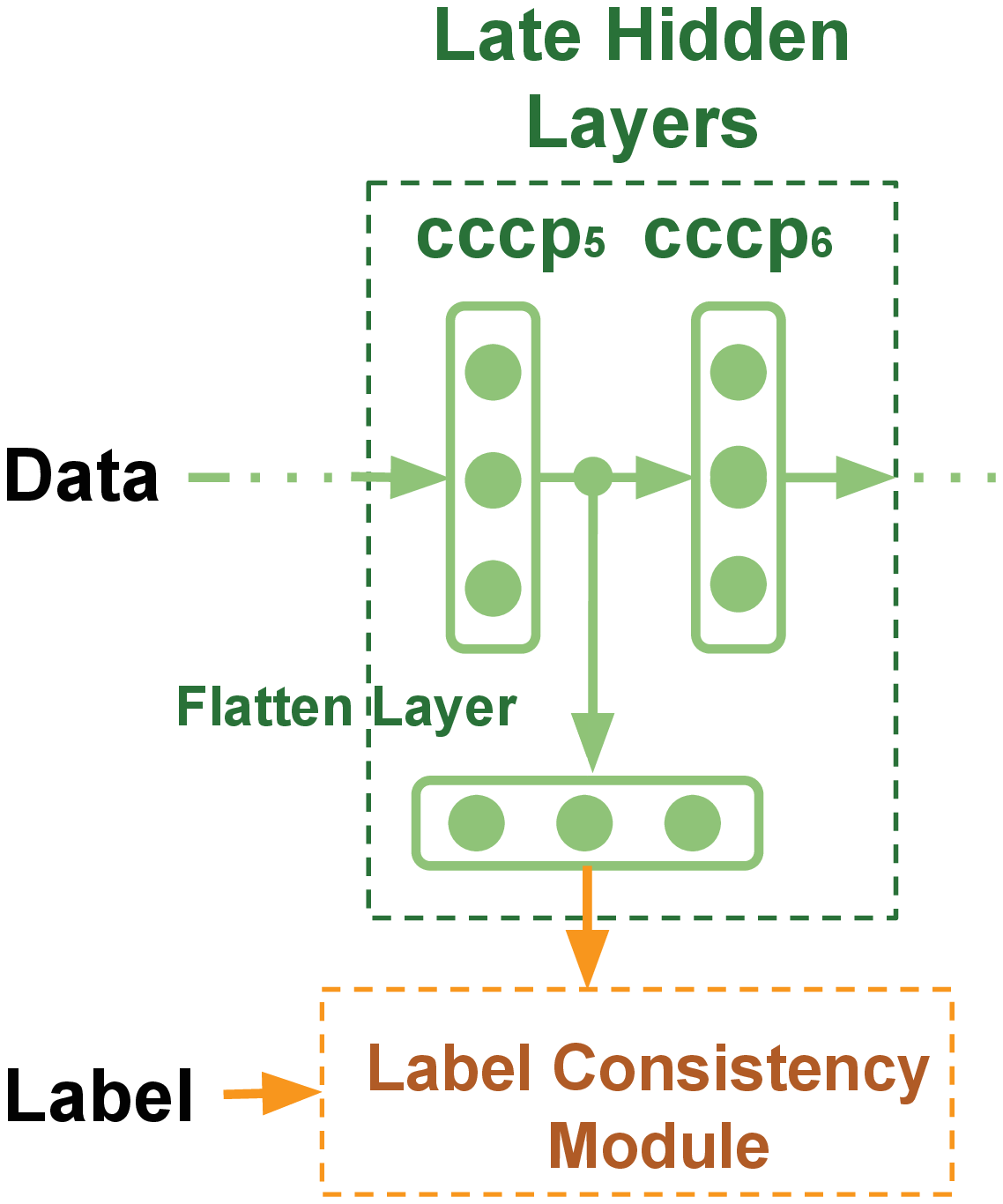}
}\\
\subfigure[]
{
\label{fig6:c}
\includegraphics[width=0.93\linewidth]{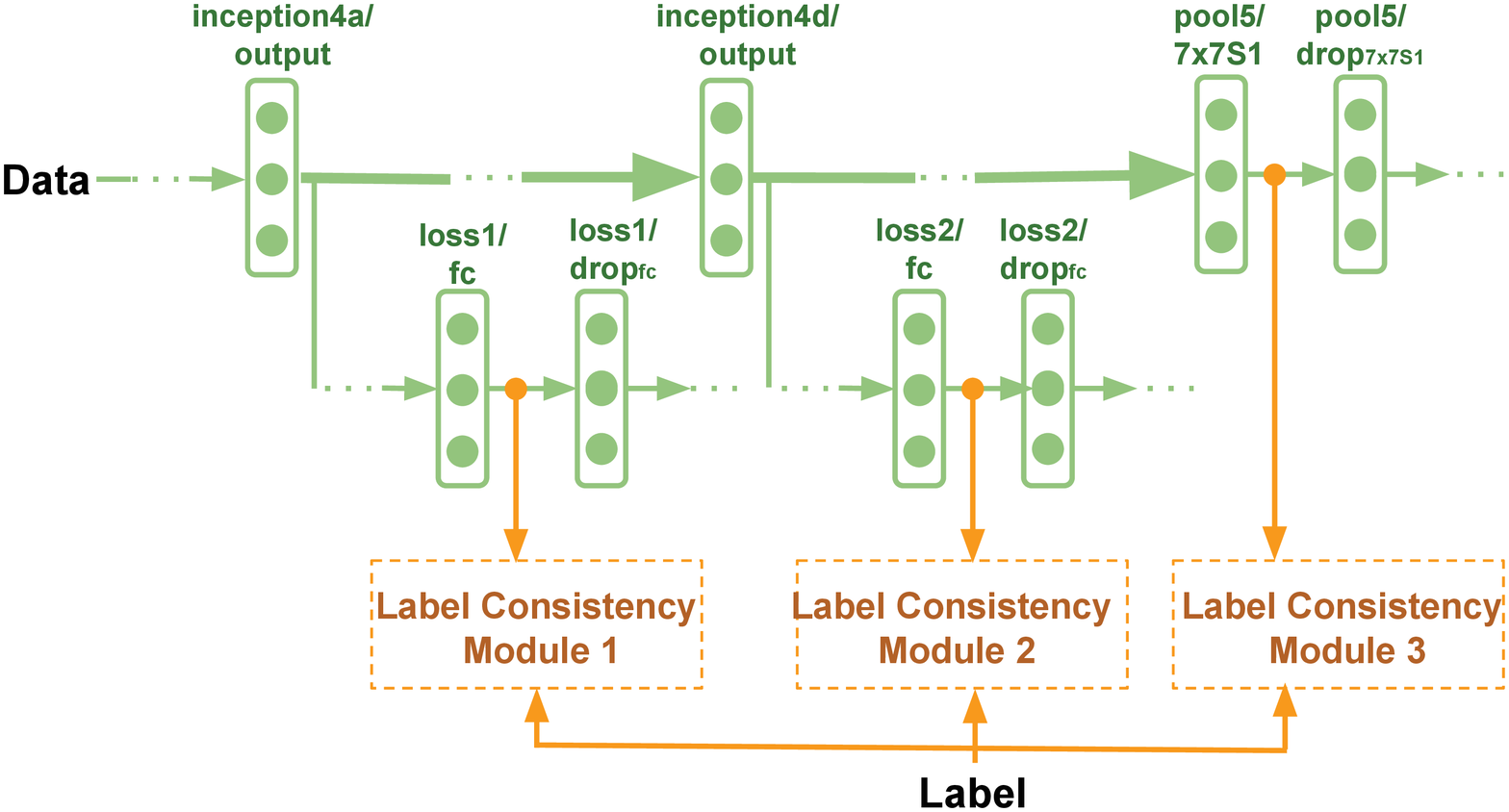}
}
\end{center}
\vspace{-0.1cm}
\caption{Examples of direct (explicit) supervision in the late hidden layers including (a) $\text{fc}_7$ layer in the CNN architectures including VGGNet~\cite{vgg} and AlexNet~\cite{hinton12}; (b) CCCP5 layer in the Network-in-Network~\cite{nin};(c) $\text{loss}_1/\text{fc}$, $\text{loss}_2/\text{fc}$
and $\text{Pool}_5/7\times7\text{S}_1$ in the GoogLeNet~\cite{googlenet}. The symbol of three dots denotes other layers in the network.}
\label{fig6}
\end{figure}

We use the VGGNet-16 architecture \cite{vgg} as in~\cite{twostream2} for two
streams where the explicit supervision is added in the late hidden layer
$\text{fc}_7$, which is the second fully-connected layer. More specifically, we feed the output of layer $\text{fc}_7$ to a fully-connected layer (denoted as $\text{fc}_{7.5}$) to produce the transformed representation, and compare it to the ideal discriminative representation
$\mathbf{q}^{(\text{fc7})}$. The implementation of this explicit supervision is shown in Figure~\ref{fig6:a}. Since UCF101 has 101 classes and the $\text{fc}_7$ layer of VGGNet has output dimension $4096$, the output of $\text{fc}_{7.5}$ has the same size $4096$, and
around 40 neurons are associated to each class. For both streams, we set $\alpha = 0.05$ in (\ref{eq10}) to balance the two loss terms.

\emph{Benefits of Adding Explicit Supervision to Late Hidden Layers.}
We aim to demonstrate the benefits of adding explicit supervision to late hidden layers. We first obtain the baseline result by running the standard two-stream CNN implementation provided by~\cite{twostream2}, which uses softmax classification loss only to train the spatial and temporal nets. Then we remove the softmax layers from this two-stream CNN but add explicit supervision to the $\text{fc}_7$ hidden layers. We call this network as `LCNN-1'. Next we maintain the softmax layers in the standard two-stream CNN but add explicit supervision to the $\text{fc}_7$ layers. We call this network as `LCNN-2'. Please note that we do use the same parameter setting and initial model in these three types of neural networks. The results are summarized in Table \ref{tb2}. It can be seen from the results of LCNN-1 that even without the help of the classifier, our label consistency constraint alone is very effective for learning discriminative features and achieves better classification performance than the baseline. We can also see that adding explicit supervision to late hidden layers not only improves the classification results at the output layer (LCNN-2 (argmax)), but also generates discriminative representations which achieve better results even with a simple $k$-NN classifier (LCNN-2 ($k$-NN)). In addition, we compare LCNN with other state-of-the-art approaches in Table~\ref{tb1}.

\emph{Discriminability of Learned Representations.}
We visualize the representations of \textit{test videos} generated by late hidden layers $\text{fc}_{7.5}$,
$\text{fc}_{7}$ and $\text{fc}_{6}$ in
Figure \ref{fig2}. It can be seen that the entries of layer $\text{fc}_{7.5}$ representations in Figure~\ref{fig2:i4} are very peaked at the corresponding class,
which forms a very good approximation to the ideal discriminative representation. Please note that a video of a testing class certainly can (and does) use neurons
from other classes as shown in Figure~\ref{fig2:i4}. It indicates that sharing features between classes is not prohibited. Further notice that such discriminative capability is
achieved during testing, which indicates that LCNN generalizes well without severe overfitting. For $\text{fc}_7$ and $\text{fc}_6$ representations in Figures~\ref{fig2:c4} and~\ref{fig2:g4},
their entropy has decreased, which means that the discriminativeness of previous layers benefits from the backpropagation of the discriminative representation error introduced by LCNN.
In Figure~\ref{fig5}, we plot the performance curves for a range of $k$ (recall $k$ is the number of nearest neighbors for a $k$-NN classifier) using LCNN-2. We observe that our approach is
insensitive to the selection of $k$, likely due to the increase of inter-class distances in generated class-specific representations.

\emph{Smaller Training and Testing Errors.}
We investigate the convergence and testing error of LCNN during network training. We plot the testing error and training error w.r.t. number of epochs from spatial net in Figure~\ref{fig3}. It can be seen that LCNN has smaller training error than the baseline (VGGNet-16), which can converge more quickly and alleviate gradient vanishing due to the explicit supervision to late hidden layers. In addition, LCNN has smaller testing error compared with the baseline, which means that LCNN has better generalization capability.

\begin{table}[t]
\centering
\begin{tabular}{|c|c|c|c|}
\hline
Network Architecture  & Spatial  & Temporal & Both \\
\hline
VGGNet-16~\cite{twostream2} & 54.5 & 42.6 & - \\
ClarifaiNet~\cite{twostream1}  & 42.3 & 47 & - \\
GoogLeNet~\cite{googlenet} &53.7 & 39.9 & - \\
baseline & 55.8 & 41.8 & - \\
LCNN-1& 56.9 & 45.1 & 59.8\\
LCNN-2 (argmax) & 57.3 & 44.9 & 61.7\\
LCNN-2 ($k$-NN) & 58.6 & 45.9 & 62.6\\
\hline
\end{tabular}
\vspace{0.2cm}
\caption{Mean Average Precision performance on the THUMOS15 validation set. The results
of~\cite{twostream2,twostream1,googlenet} are copied from~\cite{twostream2}. The `baseline' are the results of running
the two-stream CNN implementation provided by~\cite{twostream2}. LCNN and baseline are trained with the same parameter setting and initial model.
Our result $62.6\%$ mAP is also better than $54.7\%$ using method in~\cite{Lan15},
which is reported in~\cite{thumos15}.}
\label{tb3}
\end{table}

\subsubsection{THUMOS15 Dataset} \label{sec5_2_2}
Next we evaluate our approach on the more challenging THUMOS15 challenge action dataset. It includes all 13,320 video clips
from UCF101 dataset for training, and $2,104$ temporarily \textit{untrimmed} videos from the 101 classes for validation.
We employ the standard Mean Average Precision (mAP) for THUMOS15 recognition task to evaluate LCNN.

We use two-stream CNN based on VGGNet-16 discussed in Section \ref{sec5_2_1}, where explicit supervision is added in the $\text{fc}_{7}$ layers.
We train it using all UCF101 data. We used the evaluation tool provided by the dataset provider to evaluate mAP performance, which requires
the probabilities for each category for a testing video. For our two classification schemes, \textit{i.e.} argmax and $k$-NN, we
use different approaches to generate the probability prediction for a testing video. For argmax, we can directly use the
output layer. For the $k$-NN scheme, given the representation from $\text{fc}_{7.5}$ layer, we compute a sample's distances to classes only presented in its $k$
nearest neighbors, and convert them to similarity weights using a Gaussian kernel and set other classes to have very low similarity; finally we calculate the probability by doing L1 normalization on the similarity vector.

We obtained the baseline by running the two-stream CNN implementation provided by~\cite{twostream2}. We compare our LCNN results with the baseline and other state-of-the-art approaches~\cite{twostream2,twostream1,googlenet} on the THUMOS15 dataset.
The results are summarized in Table \ref{tb3}. LCNN-1 is better than the baseline and LCNN-2 can further improve the mAP performances. Our results in the spatial stream outperform the results in~\cite{twostream2},~\cite{twostream1} and~\cite{googlenet},
while our results in the temporal stream are comparable to~\cite{twostream1}. Based on this experiment, we can see that LCNN is highly effective and generalizes well to more complex testing data.

\begin{table}[t]
\centering
\begin{tabular}{|l|c|}
\hline
Method (Without Data Augment.) & Test Error (\%) \\
\hline
Stochastic Pooling \cite{stochasticpool} & 15.13 \\
Maxout Networks \cite{maxout} & 11.68 \\
DSN \cite{dsn} & 9.78 \\
baseline & 10.41 \\
LCNN-2 (argmax) & 9.75 \\
\hline
\hline
Method (With Data Augment.) & Test Error (\%) \\
\hline
Maxout Networks \cite{maxout} & 9.38 \\
DropConnect \cite{dropconnect} & 9.32 \\
DSN \cite{dsn} & 8.22 \\
baseline & 8.81 \\
LCNN-2 (argmax) & 8.14 \\
\hline
\end{tabular}
\vspace{0.2cm}
\caption{\label{tb4}Test error rates from different approaches on the CIFAR-10 dataset. The results of~\cite{stochasticpool,maxout,dropconnect,dsn} are copied from~\cite{nin}. The `baseline' is the result of Network in Network (NIN)~\cite{nin}.
Following~\cite{dsn}, LCNN-2 is also trained on top of the NIN implementation provided by~\cite{nin}. The only difference between the baseline and LCNN-2 is
that we add the explicit supervision to the $\text{cccp}_5$ layer for LCNN-2.}
\end{table}

\subsection{Object Recognition} \label{sec5_3}

\subsubsection{CIFAR-10 Dataset} \label{sec5_3_1}
The CIFAR-10 dataset contains $60,000$ color images from 10 classes, which are split into $50,000$ training images and
10,000 testing images. We compare LCNN-2 with several recently proposed techniques, especially the Deeply Supervised Net (DSN)~\cite{dsn},
which adds explicit supervision to all hidden layers. For our underlying architecture, we also choose Network in Network (NIN)~\cite{nin}
as in~\cite{dsn}. We follow the same data augmentation techniques in~\cite{nin} by zero padding on each side, then do corner cropping and
random flipping during training.

For LCNN-2, we add the explicit supervision to the $5^{\text{th}}$ cascaded cross channel parametric
pooling layer ($\text{cccp}_5$)~\cite{nin}, which is a late $1\times 1$ convolutional layer. We first flatten the
output of this convolutional layer into a one dimensional vector, and then feed it into a fully-connected layer (denoted
as $\text{fc}_{5.5}$) to obtain the transformed representation. This implementation is shown in Figure~\ref{fig6:b}. We set the hyper-parameter $\alpha =0.0375$ during training.
For classification, we adopt the argmax classification scheme.

The baseline result is from NIN~\cite{nin}. LCNN-2 is constructed on top of the NIN implementation provided by~\cite{nin} with the same parameter setting and initial model.
We compare our LCNN-2 result with the baseline and other state-of-the-art approaches including DSN~\cite{dsn}. The results are summarized in Table \ref{tb4}.
Regardless of the data augmentation, LCNN-2 consistently outperforms all previous methods, including the baseline NIN~\cite{nin} and DSN~\cite{dsn}. The results are impressive, since DSN adds
an SVM loss to every hidden layer during training, while LCNN-2 only adds a discriminative representation error loss to
one late hidden layer. It suggests that adding direct supervision to the more category-specific late hidden layers might
be more effective than to the early hidden layers which tend to be shared across categories.

\begin{table}[t]
\centering
\begin{tabular} {|c|c|c|}
\hline
Network Architecture & Top-1 (\%) & Top-5 (\%) \\
\hline
GoogLeNet~\cite{googlenet} & - & 89.93 \\
AlexNet~\cite{hinton12} & 58.9 & - \\
Clarifai~\cite{fergus14} & 62.4 & -\\
baseline & 62.64 & 85.54 \\
LCNN-2 (argmax) & 68.68 & 89.03 \\
\hline
\end{tabular}
\vspace{0.2cm}
\caption{\label{tb5}Recognition Performances using different approaches on the ImageNet 2012 Validation set. The result of~\cite{googlenet} is copied from original paper while the results of~\cite{hinton12,fergus14} are copied from~\cite{targetcoding}. The `baseline' is the result of running the GoogLeNet implementation in CAFFE toolbox. The only difference between the baseline and LCNN-2 is that we add explicit supervision to three layers ($\text{loss}_1/\text{fc}$, $\text{loss}_2/\text{fc}$ and $\text{Pool}_5/7\times7\text{S}_1$) for LCNN-2.}
\end{table}

\subsubsection{ImageNet Dataset} \label{sec5_3_2}

In this section, we demonstrate that LCNN can be combined with state-of-the-art CNN architecture
GoogLeNet~\cite{googlenet}, which is a most recent very deep CNN with 22 layers and achieved the best performance on
ILSVRC 2014. The ILSVRC classification challenge contains about 1.2 million training images and $50,000$ images for
validation from 1,000 categories.

To tackle such a very deep network architecture, we construct LCNN on top of the GoogLeNet implementation in CAFFE toolbox by adding explicit supervision to multiple late hidden layers instead of a
single one. Specifically, as shown in Figure~\ref{fig6:c}, the discriminative representation error losses are added to three layers:
$\text{loss}_1/\text{fc}$, $\text{loss}_2/\text{fc}$
and $\text{Pool}_5/7\times7\text{S}_1$ with the same weights used for the three softmax loss layers in~\cite{googlenet}. We evaluate our
approach in terms of top-1 and top-5 accuracy rate. we adopt the argmax classification scheme.

The baseline is the result of running GoogLeNet implementation in CAFFE toolbox. Our LCNN-2 and GoogLeNet are trained on the ImageNet dataset from scratch with the same parameter setting.
The results are listed in Table \ref{tb5}. LCNN-2 outperform the baseline in both
evaluation metrics with the same parameter setting. Please note that we did not get the same result reported in GoogLeNet~\cite{googlenet} by simply running the implementation in CAFFE.
Our goal here is to show that as the network becomes deeper, learning good discriminative features for
hidden layers might become more difficult solely depending on the prediction error loss. Therefore, adding explicit
supervision to late hidden layers under this scenario becomes particularly useful.

\subsubsection{Caltech101 Dataset} \label{sec5_3_3}
Caltech101 contains $9,146$ images from 101 object categories and a background category. In this experiment, we test the
performance of LCNN with a limited amount of training data, and compare it with several state-of-the-art approaches,
including label consistent K-SVD~\cite{lcksvd}.

For fair comparison with previous work, we follow the standard classification settings. During training time, 30 images
are randomly chosen from each category to form the training set, and at most 50 images per category are tested. We use
the ImageNet trained model from AlexNet in~\cite{hinton12} and VGGNet-16 in~\cite{vgg}, and fine-tune them on the Caltech101 dataset.
We built our LCNN on top of AlexNet and VGGNet-16 respectively in this experiment. The explicit supervision is added to the second fully-connected layer ($\text{fc}_7$). We set the hyperparameter $\alpha=0.0375$.

The baseline is the result of fine-tuning AlexNet on Caltech101. Then we finetune our LCNN with the same parameter setting and initial model.
Similarly, we obtained the baseline* result and LCNN results based on VGGNet-16. The results are summarized in Table \ref{tb6}. With only a limited amount of data available, our approach makes better use of the training data and achieves higher accuracy. LCNN outperforms both the baseline results and other deep learning approaches, representing state-of-the-art on this task.

\begin{table}[t]
\centering
\begin{tabular} {|l|c|}
\hline
Method & Accuracy(\%)\\
\hline
LC-KSVD~\cite{lcksvd}& 73.6 \\
\hline
Zeiler~\cite{fergus14}& 86.5 \\
Dosovitskiy~\cite{Dosovitskiy14}& 85.5 \\
Zhou~\cite{Zhou14}& 87.2 \\
He~\cite{He14} & 91.44 \\
\hline
baseline & 87.1 \\
LCNN-1 ($k$-NN) & 88.51 \\
LCNN-2 (argmax) & 90.11 \\
LCNN-2 ($k$-NN) & 89.45 \\
\hline
baseline* & 92.5 \\
LCNN-2* (argmax) & 93.7 \\
LCNN-2* ($k$-NN) & 93.6 \\
\hline
\end{tabular}
\vspace{0.2cm}
\caption{\label{tb6} Comparisons of LCNN with other approaches on the Caltech101 dataset. The results of~\cite{lcksvd,fergus14,Dosovitskiy14,Zhou14,He14} are copied from their original papers. The `baseline' and `baseline*' are the results by fine-tuning AlexNet model~\cite{hinton12} and VGGNet-16 model~\cite{vgg} on Caltech101 dataset, respectively. LCNN-1, LCNN-2 and `baseline' are trained with the same parameter setting. LCNN-2 and `baseline*' are trained with the same parameter setting as well.}
\end{table}

\section{Conclusion}\label{sec6}

We introduced the Label Consistent Neural Network, a supervised feature learning algorithm, by adding explicit
supervision to late hidden layers. By introducing a discriminative representation error and combining it with the
traditional prediction error in neural networks, we achieve better classification performance at the output layer, and
more discriminative representations at the hidden layers. Experimental results show that our
approach operates at the state-of-the-art on several publicly available action and object recognition dataset. It leads to faster
convergence speed and works well when only limited video or image data is presented. Our approach can be seamlessly combined
with various network architectures. Future work includes applying the discriminative learned category-specific representations to
other computer vision tasks besides action and object recognition.

\section*{Acknowledgement}
This work is supported by the Intelligence Advanced Research Projects Activity (IARPA) via Department of Interior National Business Center contract number D11PC20071. The U.S. Government is authorized to reproduce and distribute reprints for Government purposes notwithstanding any copyright annotation thereon. Disclaimer: The views and conclusions contained herein are those of the authors and should not be interpreted as necessarily representing the official policies or endorsements, either expressed or implied, of IARPA, DoI/NBC, or the U.S. Government.

{\small
\bibliographystyle{ieee}
\bibliography{egbib}
}

\end{document}